\documentclass[10pt,twocolumn,letterpaper]{article}

\usepackage{cvpr}              % To produce the CAMERA-READY version
\usepackage{graphicx}
\usepackage{acro} % TODO
\DeclareAcronym{rl}{
short = RL ,
long = Reinforcement Learning ,
short-plural =  ,
long-plural =  ,
}

\DeclareAcronym{mdp}{
short = MDP ,
long = Markov Decision Process ,
short-plural =  s,
long-plural =  es,
}

\DeclareAcronym{bc}{
short = BC ,
long = Behavior Cloning,
short-plural =  ,
long-plural =  ,
}

\DeclareAcronym{ml}{
short = ML ,
long = Machine Learning ,
short-plural =  ,
long-plural =  ,
}

\DeclareAcronym{mlp}{
short = MLP ,
long = Multilayer Perceptron ,
short-plural =  s,
long-plural =  s,
}

\DeclareAcronym{rnn}{
short = RNN ,
long = Recurrent Neural Network ,
short-plural = s ,
long-plural = s ,
}

\DeclareAcronym{nlp}{
short = NLP ,
long = Natural Language Processing ,
short-plural =  ,
long-plural =  ,
}

\DeclareAcronym{mse}{
short = MSE,
long = Mean Squared Error
}

\DeclareAcronym{sota}{
short = SOTA,
long = State-of-the-Art
}

\DeclareAcronym{gpt}{
short = GPT,
long = General Purpose Transformer 
}

\DeclareAcronym{cnn}{
short = CNN,
long = Convolutional Neural Network,
short-plural =  s,
long-plural =  s,
}

\DeclareAcronym{ai}{
short = AI,
long = Artifical Intelligence,
}

\DeclareAcronym{sg}{
short = SG,
long = Scene Graph,
short-plural =  s,
long-plural =  s,
}

\DeclareAcronym{sgg}{
short = SGG,
long = Scene Graph Generation,
short-plural =  s,
long-plural =  s,
}

\DeclareAcronym{gnn}{
short = GNN,
long = Graph Neural Network ,
short-plural =  s,
long-plural =  s,
}

\DeclareAcronym{adaln}{
short = AdaLN,
long = Adaptive Layer Normalization,
short-plural =  ,
long-plural =  ,
}

\DeclareAcronym{asg}{
short = ASG,
long = Affordance Scene Graph ,
short-plural =  s,
long-plural =  s,
}

\DeclareAcronym{vqa}{
short = VQA,
long = Visual Question Answering ,
short-plural =  s,
long-plural =  s,
}

\DeclareAcronym{il}{
short = IL,
long = Imitation Learning,
short-plural =  s,
long-plural =  s,
}

\DeclareAcronym{icil}{
short = ICIL,
long = In-Context Imitation Learning,
short-plural =  s,
long-plural =  s,
}

\DeclareAcronym{gcil}{
short = GCIL,
long = Goal-Conditioned Imitation Learning,
short-plural =  s,
long-plural =  s,
}

\DeclareAcronym{llm}{
short = LLM,
long = Large Language Model,
short-plural =  s,
long-plural =  s,
}

\DeclareAcronym{vlm}{
short = VLM,
long = Vision Language Model,
short-plural =  s,
long-plural =  s,
}

\DeclareAcronym{rt}{
short = RT,
long = Robotic Transformer,
short-plural =  s,
long-plural =  s,
}

\DeclareAcronym{vit}{
short = ViT,
long = Vision Transformer,
short-plural =  s,
long-plural =  s,
}

\DeclareAcronym{dot}{
short = DOT,
long = Decoder-Only Transformer,
short-plural =  s,
long-plural =  s,
}

\DeclareAcronym{gc}{
short = GC,
long = Graph Convolution,
short-plural =  s,
long-plural =  s,
}

\DeclareAcronym{beso}{
short = BESO,
long = BEhavior generation with ScOre-based Diffusion Policies,
short-plural =  ,
long-plural =  ,
}

\DeclareAcronym{mdt}{
short = MDT,
long = Multimodal Diffusion Transformer,
short-plural =  s,
long-plural =  s,
}

\DeclareAcronym{vqbet}{
short = VQ-BeT,
long = Vector-Quantized Behavior Transformer,
short-plural =  s,
long-plural =  s,
}

\DeclareAcronym{cbet}{
short = C-BeT,
long = Conditional-Behavior Transformer,
short-plural =  s,
long-plural =  s,
}

\DeclareAcronym{ibc}{
short = IBC,
long = Implicit Behavior Cloning,
short-plural =  s,
long-plural =  s,
}

\DeclareAcronym{tasg}{
short = TESA,
long = Task-Agnostic Embedding of Scene Graphs using Vision Alignment,
short-plural =  s,
long-plural =  s,
}

\DeclareAcronym{clip}{
short = CLIP,
long = Contrastive Language-Image Pretraining,
short-plural =  s,
long-plural =  s,
}

\DeclareAcronym{ir}{
short = IR,
long = Image Retrieval,
short-plural =  s,
long-plural =  s,
}

\DeclareAcronym{gqa}{
short = GQA,
long = Generalized Question Answering,
short-plural =  s,
long-plural =  s,
}

\DeclareAcronym{psg}{
short = PSG,
long = Panoptic Scene Graphs,
short-plural =  s,
long-plural =  s,
}

\DeclareAcronym{vg}{
short = VG-150,
long = Visual Genome,
short-plural =  s,
long-plural =  s,
}

\DeclareAcronym{dino}{
short = DINOv2,
long = DIstillation of Knowledge with NO labels,
short-plural =  s,
long-plural =  s,
}

\DeclareAcronym{coco}{
short = COCO,
long = Common Objects in Context,
short-plural =  s,
long-plural =  s,
}

\DeclareAcronym{vfm}{
short = VFM,
long = Vision Foundation Model,
short-plural =  s,
long-plural =  s,
}

\DeclareAcronym{span}{
short = SPAN,
long = Scene Graph-Image Contrastive Learning Framework,
short-plural =  s,
long-plural =  s,
}

\DeclareAcronym{dcr}{
short = DCR,
long = Dense Caption Reasoning,
short-plural =  s,
long-plural =  s,
}

\DeclareAcronym{lstm}{
short = LSTM,
long = Long Short-Term Memory,
short-plural =  s,
long-plural =  s,
}

\DeclareAcronym{pca}{
short = PCA,
long = Principle Component Analysis,
short-plural =  s,
long-plural =  s,
}

\DeclareAcronym{siglip}{
short = SigLIP,
long = Sigmoid Loss for Language Image Pre-Training,
short-plural =  s,
long-plural =  s,
}

\DeclareAcronym{shaun}{
short = SHAUN,
long = \textbf{S}cene Grap\textbf{H}s \textbf{A}s Str\textbf{U}ctured Represe\textbf{N}tations,
short-plural =  s,
long-plural =  s,
}

\DeclareAcronym{tamp}{
short = TAMP,
long = Task and Motion Planning,
short-plural =  s,
long-plural =  s,
}

\DeclareAcronym{mpc}{
short = MPC,
long = Model Predictive Control,
short-plural =  s,
long-plural =  s,
}

\DeclareAcronym{ebm}{
short = EBM,
long = Energy Based Model,
short-plural =  s,
long-plural =  s,
}

\DeclareAcronym{cdm}{
short = CDM,
long = Concept Distillation Module,
short-plural =  s,
long-plural =  s,
}

\DeclareAcronym{gatv2}{
short = GATv2,
long = Graph Attention v2,
short-plural =  s,
long-plural =  s,
}

\DeclareAcronym{sgat}{
short = SGAT,
long = Sparse Graph Attention,
short-plural =  s,
long-plural =  s,
}

\DeclareAcronym{sgatv2}{
short = SGATv2,
long = Sparse Graph Attention v2,
short-plural =  s,
long-plural =  s,
}

\DeclareAcronym{cgr}{
short = CGR,
long = Causal Graph Regression,
short-plural =  s,
long-plural =  s,
}

\DeclareAcronym{ibcs}{
short = IBCS,
long = Information Bottleneck-Constrained Denoised Causal Subgraph,
short-plural =  s,
long-plural =  s,
}

\DeclareAcronym{mog}{
short = MoG,
long = Mixture of Graphs,
short-plural =  ,
long-plural =  ,
}

\DeclareAcronym{diffpool}{
short = DiffPool,
long = Differentiable Pooling,
short-plural =  ,
long-plural =  ,
}

\DeclareAcronym{sagpool}{
short = SAGPool,
long = Self-Attention Graph Pooling,
short-plural =  ,
long-plural =  ,
}

\DeclareAcronym{grepool}{
short = GrePool,
long = Graph Explicit Pooling,
short-plural =  ,
long-plural =  ,
}

\DeclareAcronym{ib}{
short = IB,
long = Information Bottleneck,
short-plural =  s,
long-plural =  s,
}

\DeclareAcronym{hts}{
short = HTS,
long = Heterogeneous Task Settings,
short-plural =  s,
long-plural =  s,
}

\DeclareAcronym{grad_cam}{
short = Grad-CAM,
long = Gradient-weighted Class Activation Mapping,
short-plural =  s,
long-plural =  s,
}

\DeclareAcronym{wc}{
short = WC,
long = World Coordinates,
short-plural =  s,
long-plural =  s,
}

\DeclareAcronym{rc}{
short = RC,
long = Robot Coordinates,
short-plural =  s,
long-plural =  s,
}

\DeclareAcronym{sir}{
short = SIR,
long = \textbf{S}tructured \textbf{I}mage \textbf{R}epresentations,
short-plural =  s,
long-plural =  s,
}

\DeclareAcronym{fcg}{
short = FC-Graph,
long = Fully-Connected-Graph,
short-plural =  s,
long-plural =  s,
}

\DeclareAcronym{ns}{
short = NS,
long = node score,
short-plural =  s,
long-plural =  s,
}
\usepackage{booktabs}
\usepackage{multirow}
\usepackage{adjustbox}

\definecolor{cvprblue}{rgb}{0.21,0.49,0.74}
\usepackage[pagebackref,breaklinks,colorlinks,allcolors=cvprblue]{hyperref}
\usepackage[accsupp]{axessibility} 

\def\paperID{7398} % *** Enter the Paper ID here
\def\confName{CVPR}
\def\confYear{2026}

\title{SIR: Structured Image Representations for Explainable Robot Learning}

\author{\textbf{Paul Mattes}$^{1}$, 
  \textbf{Jan Schwab}$^{1}$, 
  \textbf{Jens Bosch}$^{1}$, 
  \textbf{Nils Blank}$^{1}$, \\
  \textbf{Maximilian Xiling Li}$^{1}$, 
  \textbf{Minh-Trung Tang}$^{1}$, 
  \textbf{Moritz Haberland}$^{1}$, 
  \textbf{Rudolf Lioutikov}$^{1,2}$ \\[0.5em]
  $^{1}$Intuitive Robots Lab, Karlsruhe Institute of Technology, Germany 
  $^{2}$Robotics Institute Germany \\[0.3em]
{\tt\small \{paul.mattes, lioutikov\}@kit.edu}
}

\begin{document}
\maketitle
\begin{abstract}
Existing robot policies based on learned visual embeddings lack explicit structure and are sensitive to visual distractions.
Thus, the representations that drive their behaviour are often opaque, making their decision-making process difficult to interpret.
To address this, we introduce \acf{sir}, a method that leverages \acp{sg} as an intermediate representation for robot policy learning.
Our approach first constructs a fully connected graph, using image-derived features as initial node representations. 
Then, a module learns to sparsify this graph end-to-end, creating a task-relevant sub-graph that is passed to the action generation model.
This process makes our model intrinsically explainable.
Evaluations on RoboCasa show that our sparse graph policies outperform image-based baselines on average with 19.5\% vs 14.81\% success rate.
Most importantly, we show that the learned sparse graphs are a powerful tool for model analysis.
By analysing when the model's sub-graph deviates from human expectation, such as by including distractor nodes or omitting key objects, we successfully uncover dataset biases, including spurious correlations and positional biases. \url{https://github.com/intuitive-robots/SIR_Model}
\end{abstract}    
\section{Introduction}
\begin{figure*}[t]
    \centering
    \includegraphics[width=0.95\linewidth]{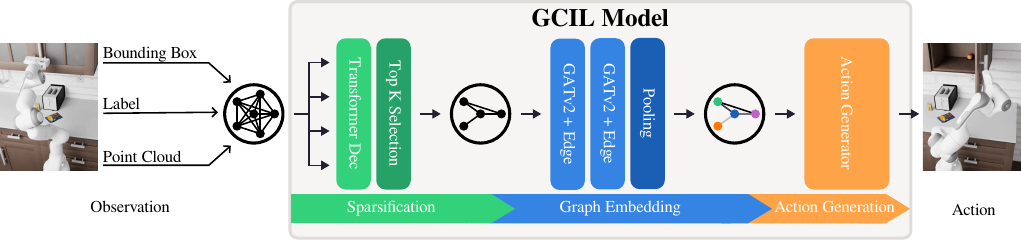}
    \caption{First, \ac{sir} extracts image features to generate an initial, fully connected \ac{sg}, using these features as node representations. This \ac{fcg} is then passed to the \ac{gcil} model. Within the model, a learnable sparsification module first prunes the \ac{fcg} into a minimal, task-relevant sub-graph, retaining only the nodes the model deems important for action generation. This sparse sub-graph is then processed by a two-layer \ac{gnn} that employs global average pooling to create a final graph embedding. This embedding serves as the state representation for the downstream action generation model.}
    \label{fig:teaser_figure}
\end{figure*}

\acf{il} \citep{ilsurvey, ilalgo} has witnessed significant advancements in robotics in recent years, primarily driven by the emergence of attention-based \citep{transformer} and diffusion-based \citep{diffusion_song} methods.
\acf{gcil} \citep{gcil} particularly benefits from these enhanced methods, enabling it to perform a wide range of tasks based on language goals \citep{bc_transformer, cbet, diffusion_x, beso, mdt}.
Concurrently, real-world robot agents need to be able to act in more complex environments with a greater number of observed objects \citep{oxe, rt2, octo, limulti}.

These developments create a growing need for structured and expressive scene representations in robot learning \citep{concept_graphs}. 
Most existing approaches rely on learned visual embeddings, often extracted from convolutional backbones or vision foundation models \cite{mdt, rt2, octo, flower}.
While such embeddings provide a compact encoding of visual information, they remain opaque and lack explicit structure. 
This makes them difficult to interpret, offering no clear explainability with respect to a generated decision.
We propose \textbf{\acf{sir} for Explainable Robot Learning}, which addresses this challenge by leveraging \acfp{sg} to introduce a structured image representation usable in \acf{gcil}.
% \acp{sg} explicitly encode objects, their attributes, and their relations in a graph. 
\acp{sg} provide a unifying representation that can incorporate diverse modalities extracted from perception, including symbolic information (e.g., object labels), geometric cues (e.g., bounding boxes and point clouds), and high-level image features.
By capturing the environment in this structured, relational form, \acp{sg} present a data structure that make robot behaviour more interpretable than image-based approaches. 
\ac{sir} further enhances the explainability of \acp{sg} by operating on a learned sparsified graph.
This sparsification process creates a minimal, task-relevant sub-graph, offering a clear insight into exactly which objects and interactions the model considers critical for its decision.
\acp{sg} generated this way thus offer a compact, expressive, and highly interpretable intermediate state representation.

Our \ac{gcil} model, \ac{sir}, first generates a fully connected \ac{sg} of the scene. A trainable module then sparsifies this graph into a sub-graph by keeping only the nodes the model considers important for the task. Finally, a two-layer \ac{gnn} embeds this sparse sub-graph to produce the final scene representation, which the action generation model uses as input.
To showcase the effectiveness of \acp{sir} representations, we evaluate two action generators: \ac{mdt} \cite{mdt}, which provides SOTA performance and adaptability, and a \ac{bc}-Transformer as a baseline.
The overall method can be seen in \Cref{fig:teaser_figure}.
The contributions of this paper are twofold:
\begin{itemize}
    \item We conduct a thorough investigation into how \acp{sg} can be used as an effective scene representation for robot learning. As part of this, we systematically analyse which image modalities serve as effective initial node representations.
    \item We propose a method for learnable \ac{sg} sparsification and analyse the resulting sub-graphs to provide insights into the model's decision-making process. This analysis allows us to investigate how the model perceives the scene by examining whether it correctly selects task-relevant objects and whether it includes irrelevant ones.
\end{itemize}
Based on these observations, we formulate four research questions: \\
\textbf{RQ1}: How do \ac{sg} embeddings compare to image embeddings as model input in \ac{gcil}? \\
\textbf{RQ2}: How does the choice of the initial node representation (e.g., symbolic labels, geometric cues, or visual features) affect the performance of \ac{sg}-based models? \\
\textbf{RQ3}: How robust are models that use \ac{sg} embeddings compared to image embeddings in regard to distractor objects? \\
\textbf{RQ4}: Do sparse \acp{sg} facilitate the interpretability and analysis of the decision-making process of the respective behaviour model? \\
We evaluate \ac{sir} in RoboCasa \citep{robocasa2024,robosuite2020} and CALVIN \cite{calvin}, a language conditioned imitation learning benchmark for long-horizon manipulation.
Our results demonstrate that policies using \acp{fcg} as input already achieve performance comparable or superior to image-based models.
Furthermore, our learnable sparsification method not only enhances explainability but also yields an additional improvement in overall performance.
We also find that graph-based representations are significantly more robust to distractor objects introduced during inference. 
Finally, the resulting sparse graphs provide a direct mechanism for the interpretability of the model behaviour, enabling analysis based on the specific nodes the model includes or excludes.
Critically, \ac{sir} is not a post-hoc explainability method, but intrinsically explainable because the sparse graph serves as a learned intermediate representation during action generation. 
This characteristic offers significantly greater potential for interpretability and analysis of the model's decision-making process. 
\section{Related Work}
Current graph-based \ac{il} methods are often plan-based \citep{symbolic_sg, gnn_efficient_interpret_robot_man, one_il_pick_place_graph, 3d_graph_mpc} or treat graphs as auxiliary input \citep{graph_bases_rep_model_with_rl, sg_embodied_navigation}.
We instead investigate how structured scene representations, in the form of \acp{sg}, can serve as the direct state for robot learning.
To our knowledge, \ac{sir} is the first approach to enable \ac{gcil} to solve complex, everyday kitchen tasks using \acp{sg} as the direct observation representation, without relying on planning.
This contrasts with recent methods, like Instant-Policy \citep{instantpolicy}, which uses graphs within a diffusion framework but is restricted to point-cloud embeddings. 
\textit{Compose by Focus} \citep{compose_sgs} uses \acp{sg} as direct input, but only for simple 3-4 node graphs, relies solely on point cloud values, and is evaluated on simple manipulation tasks.
In contrast, \ac{sir} works on larger, fully-connected or reduced \acp{sg}, integrates diverse image modalities as initial node representations, and is validated on complex everyday kitchen tasks.

% \subsection{Goal-Conditioned Imitation Learning}
% \ac{gcil} \cite{gcil} defines a specific goal that should be executed by the robot policy. 
% This goal can either be in natural language, one-hot encoded or a future image state. 
% Most commonly, natural language is used to guide the policy \cite{mdt, rt2, octo}.
% Current robot policy methods are mainly based on the Transformer \cite{transformer} architecture \cite{mdt, rt2, bc_transformer, beso} and use either \ac{bc} \cite{bc_transformer, rt2} or Diffusion \cite{mdt, beso} for action generation.

\subsection{Graph-Based Imitation Learning through Planning}
Graphs have been used for 3D object prediction, e.g. the robot has to push a block into a desired position \cite{3d_graph_mpc}.
More challenging tasks include real world pick and place, where a combination of symbolic and geometric image-based graphs are used \cite{symbolic_sg}.
The symbolic graph is grounded through the geometric graph.
Furthermore, graphs can include object and goal nodes to predict which object interaction is needed to achieve a specific goal \cite{gnn_efficient_interpret_robot_man, one_il_pick_place_graph}.
Another approach, called ConceptGraphs \cite{concept_graphs}, builds 3D \acp{sg} to input into an \ac{llm} for downstream plan-based task execution.
It appears that many graph-based \ac{il} approaches in robotics involve some sort of planning algorithm, like \ac{tamp} or something similar \cite{3d_graph_mpc, symbolic_sg, gnn_efficient_interpret_robot_man, one_il_pick_place_graph, concept_graphs}.
The plan-based approaches use high-level information graphs, that need specific key-points to update the graphs during execution.
Otherwise, their internal information would not change.
These key-points are much easier to identify using planning algorithms.
\ac{sir} demonstrates that graphs work for step-based \ac{gcil}, using low-level information graphs without the need of plan-based algorithms or specific key-points.

\subsection{Graph-Based Step-Wise Imitation Learning}
Step-wise \ac{il} using graphs is an under-researched field.
Three methods use graphs for visual imitation, where hand movement is mapped to a graph for grasping and reaching improvements \cite{graph_visual_imitation, graphirl, huang2024virl}, but using fully-connected graphs.
Another approach uses graphs for swarm movement to learn the underlying interaction mechanism \cite{cbc}. 
A third approach, Instant Policy, uses graphs for \ac{icil} in everyday robotic tasks \cite{instantpolicy}, modelling it as a graph generation problem with a learned diffusion process.
The fourth approach, called \textit{Compose by Focus}, uses \acp{sg} as a direct scene representation and input to a diffusion model, focusing on simple graphs (3-4 nodes) with point cloud features as node embeddings \cite{compose_sgs}.
In contrast, \ac{sir} does not use the graph as a direct part of behaviour learning. 
Instead, we use the graph as an intermediate representation to build a more structured scene understanding. 
Our work is the first to systematically investigate how \acp{sg} perform when using different image-derived modalities as initial node features. 
Furthermore, \ac{sir} is trained to predict the important sub-graph, thereby learning an explainable and sparse representation, which can be used for further downstream analysis.

\subsection{Learning Graph Sparsification}
Large graphs are reduced for various reasons, including storage~\cite{xu_learning_2025}, runtime~\cite{zhang_graph_2024}, out-of-distribution handling~\cite{yin_recipe_2025}, or interpretability~\cite{zheng_robust_2020}. 
These \textit{graph reduction} techniques are categorized as \textit{sparsification}, \textit{condensation} or \textit{coarsening}~\cite{hashemi_comprehensive_2024}. Condensation and coarsening are not suitable for interpretability, as the resulting graph's relation to the original is unclear.
Sparsification methods typically focus on removing edges, not nodes~\cite{ye_sparse_2023, rathee_learnt_2021, zheng_robust_2020}. 
This is inadequate for interpretability in graph classification/regression tasks. 
Removing only edges, especially in fully-connected graphs, does not significantly impact \ac{gnn} performance compared to removing nodes~\cite{rathee_learnt_2021}.
While some \ac{gnn} pooling layers (e.g., \ac{diffpool}~\cite{ying_hierarchical_2018}, \ac{grepool}~\cite{liu_careful_2023}, \ac{sagpool}~\cite{lee_self-attention_2019}) select nodes, they do so after or between message passing layers.
This is not truly interpretable, as information from unselected nodes may have already propagated into the final graph embedding.
Therefore, we cannot be certain which nodes contributed most significantly.
To the best of our knowledge, this work is the first to learn end-to-end node removal before message passing, ensuring that removed nodes have no impact on the graph embedding.

\section{Method}
The architecture of \ac{sir} is  illustrated in \Cref{fig:teaser_figure}. It includes four core components: (1) \ac{sg} extraction from a given image, (2) graph sparsification, (3) graph embedding generation and (4) action generation.
The resulting graph embedding serves as the state representation for a downstream action generation model, which, in this work, is instantiated as either \ac{mdt} \cite{mdt} or a \ac{bc}-Transformer \cite{bc_transformer}.
The three modules inside the \ac{gcil} model are trained end-to-end, while the scene graph extraction step is frozen.

\subsection{Scene Graph Generation}
We construct the initial \acp{sg} by extracting all objects in a scene using either ground-truth or predicted segmentation masks.
These objects are then constituted as the nodes in a \ac{fcg}, where
the initial node representations are derived from the given RGB or RGB-D image.
We investigate four primary feature modalities:
\begin{itemize}
    \item \textbf{Label:} The object label as a one-hot encoding.
    \item \textbf{Cropped-Image-Feature:} A visual embedding from a pre-trained network, generated from the BB crop.
    \item \textbf{BB-Coordinates:} The 2D bounding box corner and centre coordinates, normalized in pixel space.
    \item \textbf{Point-Cloud-Feature:} An embedding from a pre-trained network, applied to the object's associated point cloud.
\end{itemize}
A key aspect is that these feature modalities can be easily concatenated as the initial node embedding in the graph itself by design.
When including either bounding box or point cloud information, the edge features between nodes are initialised with geometric distance. 
Otherwise, the edge features are initialised with 1, in order to aid the message passing in the later graph embedding stage.

\subsection{Scene Graph Sparsification}
Although graphs provide the scene information in a more structured way than images, a \ac{fcg} contains all nodes available in the scene, reducing its interpretability.
We therefore aim to extract the most relevant sub-graph for a specific task.
Since the information in this extracted sub-graph is the only information about the scene available to the action generation model, this sub-graph serves as the explanation for the generated actions. 
To extract task-relevant \acp{sg}, \ac{sir} calculates a score for every node and uses the highest scored nodes of a graph. 
To predict these \acp{ns}, we use a two-layer Transformer-Decoder \cite{transformer} architecture with four heads per layer.
Every layer employs \ac{adaln} \cite{perez_film_2018}, which conditions the node embeddings on the language goal. 
We will refer to this module as \textit{FiLMDecoder}. 
We define the resulting node weight (NW) as
\begin{equation}
    \text{NW}(n) = \begin{cases}
        \text{NS}(n), & \text{if n is selected in the sub-graph.} \\
        0, & \text{otherwise,}        
    \end{cases}
\end{equation}
for a node $n$ based on the score $\text{NS}(n)$ predicted by the FiLMDecoder. 
To prevent a collapse where all node scores converge to a similar value, we introduce a \textit{soft histogram loss}. 
This loss encourages the predicted scores to be uniformly distributed in the [0,1] range.
Instead of hard binning, our method employs a Gaussian kernel to softly assign each score to multiple histogram bins.
These assignments are summed to form a differentiable soft histogram, which is then normalized.
Finally, we compute the \ac{mse} between this soft histogram and a uniform target distribution. 
During training, we apply a weight of 0.1 to this loss term.
The nodes with the highest scores are then chosen for the sub-graph using instruction-grounded node selection. Specifically, we select the top-k highest-scoring nodes from the graph, where k is a task-specific parameter chosen according to the number of task-relevant objects. Additionally, the model is guided with an additional L1 loss applied to the node weights, encouraging NW(n) to be high for instruction-relevant nodes and low for irrelevant ones.

\subsection{Scene Graph Embedding Generation}
We generate the \ac{sg} embedding using a \ac{gnn} composed of two \ac{gatv2} residual layers, followed by a global average pooling layer.
The \ac{gatv2} layers update node features by weighting information from neighbours using learned attention scores.
The final graph-level representation is obtained by averaging the features of all nodes after the propagation steps.
Consequently, this global average pooling aggregates features from all nodes into the final graph embedding, including those that may have received low attention weights during the graph propagation.

\paragraph{Differentiability} Further, we adapt \ac{gatv2} to include edge weights.
The message passed along an edge is not only multiplied by the attention score but also by the edge weight.
To enable learning the FiLMDecoder end-to-end, we (1) keep the gradient of $\text{NS}(n)$ as the gradient of $\text{NW}(n)$, (2) include the node weights in the edge weights $\text{EdgeWeight}(u, v) = \text{NW}(u) \cdot \text{NW}(v)$, and (3) include the node scores in the pooling step. 
Point (2) ensures that during message passing, no information leaves the "removed" nodes.
Point (3) ensures that during the pooling step, no information is passed into the graph embedding, while explicitly including $\text{NS}(n)$ improves gradient flow and ensures improved learning of the FiLMDecoder:
\begin{equation}
    GraphEmbedding = \frac{\sum_{n \in V}{\text{NW}(n) * X_n}}{\sum_{n \in V}{1_{\text{NW}(n) > 0}}},
\end{equation}
where $X_n$ describes the final node feature of node $n$.
Regarding the pooled features, this is equal to mean-pooling over the kept nodes with $\text{NW}(n) > 0$. 

\subsection{Action Generation}
Actions are generated using the given downstream action generation model, which gets as input the embedded \ac{sg} and an embedded language goal using \acs{clip} \cite{clip}.
Both action generator models, \ac{mdt} and \ac{bc}-Transformer, use one observation to generate 10 future actions.
\begin{table*}[t]
\caption{The results represent success rate in percent on RoboCasa \cite{robocasa2024} over 100 rollouts with 2 model versions trained on different seeds. The numbers behind the task category names indicate the number of tasks per category. Graph Methods use Cropped-Image-Features and BB-Coordinates as initial node representations.}
\label{tab:results}
\centering
\small
\resizebox{\linewidth}{!}{
\begin{tabular}{@{}lcccccccc@{}}
\toprule
Observation & Pick/Place (8) & Doors (4) & Drawers (2) & Knobs (2) & Levers (3) & Buttons (3) & Insert (2) & \textbf{Avg (24)} \\
\midrule
Image (Baseline)                    & \textbf{1.19 $\pm$ 0.44} & 25.13 $\pm$ 0.88 & \textbf{49.75 $\pm$ 0.75} & 7.25 $\pm$ 3.75 & 23.67 $\pm$ 0.34 & 17.00 $\pm$ 0.33 & \textbf{4.75 $\pm$ 0.25} & 14.81 $\pm$ 0.02 \\
%\midrule[0.5pt]
\acl{fcg}            & 0.06 $\pm$ 0.03 & 28.62 $\pm$ 4.25 & 39.25 $\pm$ 1.75 & 14.00 $\pm$ 0.50 & 40.00 $\pm$ 2.67 & 18.83 $\pm$ 0.17 & \textbf{4.75 $\pm$ 0.75} & 16.98 $\pm$ 0.85 \\
\textbf{\ac{sir}  } (Ours)               & 0.13 $\pm$ 0.00 & \textbf{30.25 $\pm$ 0.25} & 46.25 $\pm$ 1.75 & \textbf{16.50 $\pm$ 0.00} & \textbf{48.50 $\pm$ 2.17} & \textbf{21.83 $\pm$ 1.84} & \textbf{4.75 $\pm$ 2.25} & \textbf{19.50 $\pm$ 0.33} \\
\bottomrule
\end{tabular}
}
\end{table*}

\section{Experiments}
We conduct experiments on the RoboCasa \citep{robocasa2024} and CALVIN \citep{calvin} benchmarks with two distinct action generation models: \ac{mdt} \cite{mdt} and a \ac{bc}-Transformer \cite{bc_transformer}.
For each benchmark, we trained each model configuration twice with different seeds and perform evaluation over 100 rollouts.
The 24 atomic tasks of RoboCasa \cite{robocasa2024} are evaluated using the standard groups: \textit{Pick and Place}, \textit{Drawers}, \textit{Doors}, \textit{Buttons}, \textit{Levers}, \textit{Knobs}, and \textit{Insertion}, as well as the overall 24-task average. 
For the CALVIN benchmark, we evaluate on the D → D setting.
To ensure a fair and controlled comparison between graph-based and baseline methods, we do not use the full sensory information described in RoboCasa or CALVIN.
Instead, we focus on a subset of information provided by the static cameras, to avoid biasing results with too many different input features.

\paragraph{Baselines} We evaluate one main baseline method, which incorporates images as observation input.
Four additional baselines used in ablations leverage images, point clouds or both as observation features.
All five methods are directly comparable to the graph-based representation of using either image or point cloud features.
Baseline models using images as observation input use a pre-trained ResNet18 \cite{resnet} to generate embeddings, which is fine-tuned during training.
The model denoted with "Own pre-trained Image" uses the image encoder, which was trained for embedding the cropped bounding boxes for the graph node features and is also fine-tuned during training.
Baseline models using point cloud observations input it patch-wise into the action generation model as described in FPV-Net \cite{pointcloud_emb}.
Evaluations using the same embedding network for point clouds as we did for the initial node features of our graphs resulted in worse performance.
Therefore, we kept the better results of the baseline.
More detailed information can be found in \Cref{app:pre_trained_models}.

\begin{figure}[t]
    \centering
    \includegraphics[width=0.9\linewidth]{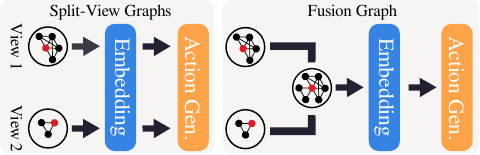}
    \caption{Multiple views of a scene observation can be handled through single embeddings using the Split-View approach or merging node features in a single graph to get one embedding for all views.}
    \label{fig:fusion}
\end{figure}

\paragraph{Graph-based observations} For RoboCasa-based experiments, graphs are generated from the static left and right camera observations. 
We explore two processing strategies: embedding them separately using two \acp{gnn} (Split-View Graph), or concatenating node features from both views into a single "Fusion Graph" processed by one \ac{gnn}.
Both approaches are shown in \Cref{fig:fusion} and can be adapted to any multi-view benchmark using static cameras. 
In contrast, the CALVIN environment only includes one static camera, thus resulting in a single graph.
\ac{sir} can either use the extracted \ac{fcg} of the image or the sparsified sub-graph to generate the graph embeddings.
Initial node features for the \acp{sg} are generated using the following methods:
The Cropped-Image-Feature is generated by first cropping the object's bounding box from the image.
This crop is then encoded using a ResNet18 backbone, which was pre-trained on an BB-image reconstruction task.
Similarly, the Point-Cloud-Feature is embedded using a pre-trained PointNet \cite{pointnet} architecture.
This network was trained on an object point cloud reconstruction task, using a Chamfer distance loss.
Label information is one-hot encoded and depends on the maximum number of objects in the scene.
BB-Coordinates are a normalized vector based on pixel-space including all four corner coordinates and the centre point.
Initial node features for all sparse graph methods are Cropped-Image-Feature and BB-Coordinates, as they achieve the highest performance, while maintaining fast inference speed.

\paragraph{Explainability Evaluation}

We evaluate our explanation subgraphs qualitatively. During the rollout, we monitor the subgraphs \(G_{sub, i}\) extracted by our sparsification method at each time step \(i\). We then construct the explanation \(G_{expl} = (V_{G_{expl}}, E_{G_{expl}})\) for a rollout (i.e., "rollout-explanation") as the mean of the extracted subgraphs. Therefore, we calculate the percentage \(p_{p, n}\) that a node (edge) \(n\) was present in the subgraph as 

\begin{equation}
    p_{p, n} = \frac{\text{Number of subgraphs \(n\) was present in}}{\text{Number of scene graphs \(n\) was present in}}.
    \label{eq:ExplanationGraphGeneration}
\end{equation}
The more all \(p_{p, n}\) (for $n \in V_{G_{expl}} \cup E_{G_{expl}}$) converge to \(\{0, 1\}\), the more consistent is the explanation throughout the rollout. The same way, we can construct an explanation subgraph for a task (i.e., "task-explanation"), by considering all time steps of all rollouts of the task. 

% To assess how explainable our representations are, we introduce the \textit{explanation graph consistency} metric. Intuitively, we calculate the percentage $p_{p, n}$ of how often a node $n$ was present in the sparsified sub-graph for every timestep with mean $\mu$ and standard deviation $\sigma$.
% This is based on the assumption that the appearance of a scene does not change significantly during a single rollout and that the same nodes are important during this rollout. 
% We thus propose the \textit{explanation graph consistency} 
% \begin{equation}
%     C_{expl} = \sum_{n \in V}{1_{p_{p, n} \in [0, p_0]}} + \sum_{n \in V}{1_{p_{p, n} \in [p_1, 1]}},
% \label{eq:ex_eval}    
% \end{equation}
% with $p_0 = max(\mu - \sigma, 0.02)$ and $p_1 = min(\mu + \sigma, 0.98)$, which is the percentage of nodes, which are present in the sub-graph with a percentage outside the standard deviation or very close to 0 or 1. 
% With increasing consistency of the sub-graph, the value of $C_{expl}$ reaches from 0 to 1.

\section{Results and Discussion}
Selected results for the evaluation of the baselines and graph-based models on RoboCasa using \ac{mdt} are displayed in \Cref{tab:results}.
All graph-based models use Cropped-Image-Features and BB-Coordinates as initial node embeddings, as well as distance based edge features.
Ablations with other initial node embeddings are shown in \Cref{tab:ablations}.
The full list of single tasks per category can be found in \Cref{app:per_task_results}.
Results for the \ac{bc} model, using generated segmentation masks and CALVIN can be found in \cref{app:bc_results}, \cref{app:gen_results} and \cref{app:calvin_results}.

\subsection{Scene Graphs as Observations}
In RoboCasa, image-based models achieve a 14.81\% average success rate, as shown in \Cref{tab:results}.
% \begin{itemize}
%     \item \textbf{Threshold:} This method selects all nodes with a score above a pre-defined, non-learned threshold hyperparameter. An additional L1 loss is used to penalize the sum of node scores, encouraging overall sparsity.
%     \item \textbf{TopK:} This method selects the sub-graph by taking the k nodes with the highest sc
\ac{fcg}-based models, without sparsification, reach nearly 17\%, and \ac{sir} with instruction-grounded sparsification widens this gap, achieving 19.5\%. Additional results in \Cref{tab:ablations} confirm this performance gap, even when using different observation types, such as point clouds or their combination with images.
However, the performance gains are not uniform across all task categories. 
While graph-based models show significant improvements in most settings (e.g., \textit{Doors}, \textit{Levers}, \textit{Knobs}, and \textit{Buttons}), they do not surpass the image-based baseline in the \textit{Drawers} or \textit{Pick and Place} task. We assume this is due to heavy dataset biases present for some tasks, as investigated in \Cref{sec:explainabiliy}.
These observations, together with results from \cref{app:bc_results}, answer \textbf{RQ1}: Graph-embeddings outperform image-embeddings across diverse task settings.

\subsection{Ablations}
Ablation results of initial graph node representations is detailed in \Cref{tab:ablations}. 
We compare the graph-based models to baselines using corresponding input features, where applicable.
Our results show that the best-performing node representations are Cropped-Image-Features, either alone (16.65\%) or in combination with BB-Coordinates (15.90\% or 16.98\% for the Fusion Graph). 
These methods outperform the standard image baseline (14.81\%) and the FiLM-conditioned image baseline on the language goals (15.85\%).
A significant gap is evident when using point cloud data. 
The baseline model using only point clouds achieves just 4.13\%, and 13.25\% when combined with images. 
In contrast, the graph-based counterparts are far more effective, reaching 11.08\% for Point-Cloud-Features alone and 15.04\% for the combination.
These observations indicate that \acp{gnn} are a more efficient architecture for integrating point cloud information as node features compared to inputting it directly into the action generation model \cite{pointcloud_emb}.
Therefore, graph-based models can effectively integrate diverse node representations and, in doing so, outperform their corresponding baselines, answering \textbf{RQ2}.

We further compare our instruction-guided sparsification method to simpler sparsification methods in \Cref{tab:sparse_ablations}. For random node removal, we simply remove random nodes from the graph. In Naive NR, node removal is learned without the soft histogram loss. We further compare \ac{sir} to Threshold node removal, where all nodes with a score over a specified threshold are retained. In TopK, we do not employ a task-specific k and do not guide the node weights with an L1 loss.
Overall, \ac{sir} with instruction-grounded sparsification outperforms other sparsification methods, whereas the soft-histogram loss has the highest impact on performance.
% In both cases, performance drops heavily compared to \ac{sir}, indicating the need for the soft histogram loss.

\begin{table}[t]
\caption{Ablation results on RoboCasa \cite{robocasa2024} over 100 rollouts with 2 model versions trained on different seeds. Ablations consider different sparsification approaches.}
\label{tab:sparse_ablations}
\centering
\begin{tabular}{@{}lc@{}}
\toprule
Sparsification Method & Avg (24) \\
\midrule[0.5pt]
None (Fully-Connected) & 16.98 $\pm$ 0.85 \\
\midrule[0.5pt]
Random Node Removal & 5.48 $\pm$ 0.19 \\
Naive NR (\textbf{no} soft histogram loss) & 9.60 $\pm$ 1.73 \\
Threshold & 17.17 $\pm$ 0.38 \\
TopK & 18.44 $\pm$ 0.77 \\
\textbf{\ac{sir}} & \textbf{19.50 $\pm$ 0.33} \\
\bottomrule
\end{tabular}
\end{table}

\begin{table}[t]
\caption{Ablation results on RoboCasa \cite{robocasa2024} over 100 rollouts with 2 model versions trained on different seeds. Ablations consider different image features used as input for the action generation model or as initial node features for the \acp{sg}.}
\label{tab:ablations}
\centering
\begin{tabular}{@{}lc@{}}
\toprule
Observation & Avg (24) \\
\midrule[0.5pt]
\textit{Baselines} &  \\
Image & 14.81 $\pm$ 0.02 \\
Image + FiLM & 15.85 $\pm$ 0.29 \\
Own Pretrained Image & 10.11 $\pm$ 0.76 \\
Point Clouds & 4.13 $\pm$ 0.13 \\
Image + Point Clouds & 13.25 $\pm$ 0.15 \\
\midrule[0.5pt]
\textit{Split-View Graph - Fully-Connected} & \\
Cropped-Img & 16.65 $\pm$ 0.23 \\
BB-Coordinates + Label & 10.98 $\pm$ 0.44 \\
BB-Coordinates + Cropped-Img & 15.90 $\pm$ 0.32 \\
Point Clouds & 11.08 $\pm$ 0.29 \\
Label + Point Clouds & 12.50 $\pm$ 1.0 \\
Cropped-Img + Point Clouds & 15.04 $\pm$ 0.09 \\
\midrule[0.5pt]
\textit{Fusion Graph - Fully-Connected} & \\
BB-Coordinates + Label & 11.06 $\pm$ 0.57 \\
BB-Coordinates + Cropped-Img & 16.98 $\pm$ 0.85 \\
\bottomrule
\end{tabular}
\end{table}

\begin{figure*}[t]
    \centering
    \includegraphics[width=0.9\linewidth]{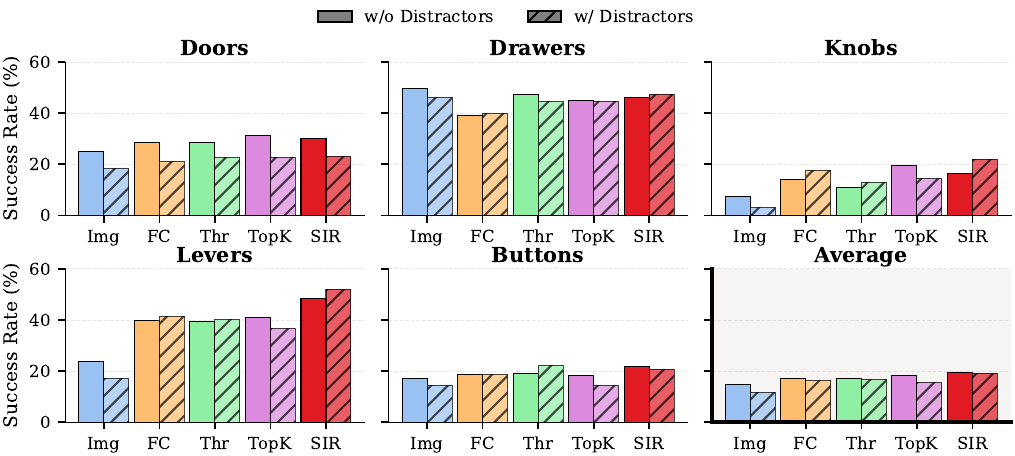}
    \caption{When novel distractor objects, not seen during training, are introduced at inference, the image-based baseline's (Img) performance drops significantly. A similar decrease is seen in our TopK sparsified model. In sharp contrast, the fully-connected (FC), Threshold (Thr) and the \ac{sir} graph models demonstrate high robustness, maintaining their average performance. These graph-based models even show performance increases in specific categories, such as Knobs and Levers. More details in \Cref{tab:distractor_tasks} in the Appendix.}
    \label{fig:distractor}
\end{figure*}

\subsection{Distractor Objects}
We further evaluate the robustness of \acp{sg} compared to images with respect to multiple distractor objects present in the scene. 
In particular, we place between 3 and 9 additional objects in the environment. 
The results are displayed in \Cref{fig:distractor}, with more detailed results in \Cref{tab:distractor_tasks} in the Appendix.
Including distractor objects results in a clear performance decrease for the image baseline, which drops by 3.3\% on average. 
A similar drop of 2.9\% is seen for our model using TopK node removal.
In contrast, \ac{sir}, the \ac{fcg}-based model and the Threshold model show almost no performance degradation on average when distractor objects are introduced.
In fact, these models even show slight performance increases in some task settings, such as \textit{Drawers}, \textit{Knobs}, and \textit{Levers}. 
However, all models, including the graph-based ones, decrease in performance on the \textit{Doors} task.
This provides a clear answer to \textbf{RQ3}: Image-based models are sensitive to novel distractor objects, whereas \ac{sir}, \ac{fcg} models and Threshold models are highly robust, achieving similar performance as before.

\begin{figure}[t]
    \centering
    \includegraphics[width=0.9\linewidth]{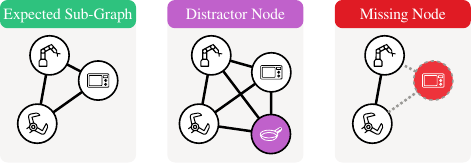}
    \caption{Possible sub-graph generation, when using our explanation graph consistency metric.}
    \label{fig:subgraphs}
\end{figure}

\begin{figure*}[t]
    \centering
    \begin{subfigure}[t]{0.45\textwidth}
        \centering
        \includegraphics[width=\textwidth]{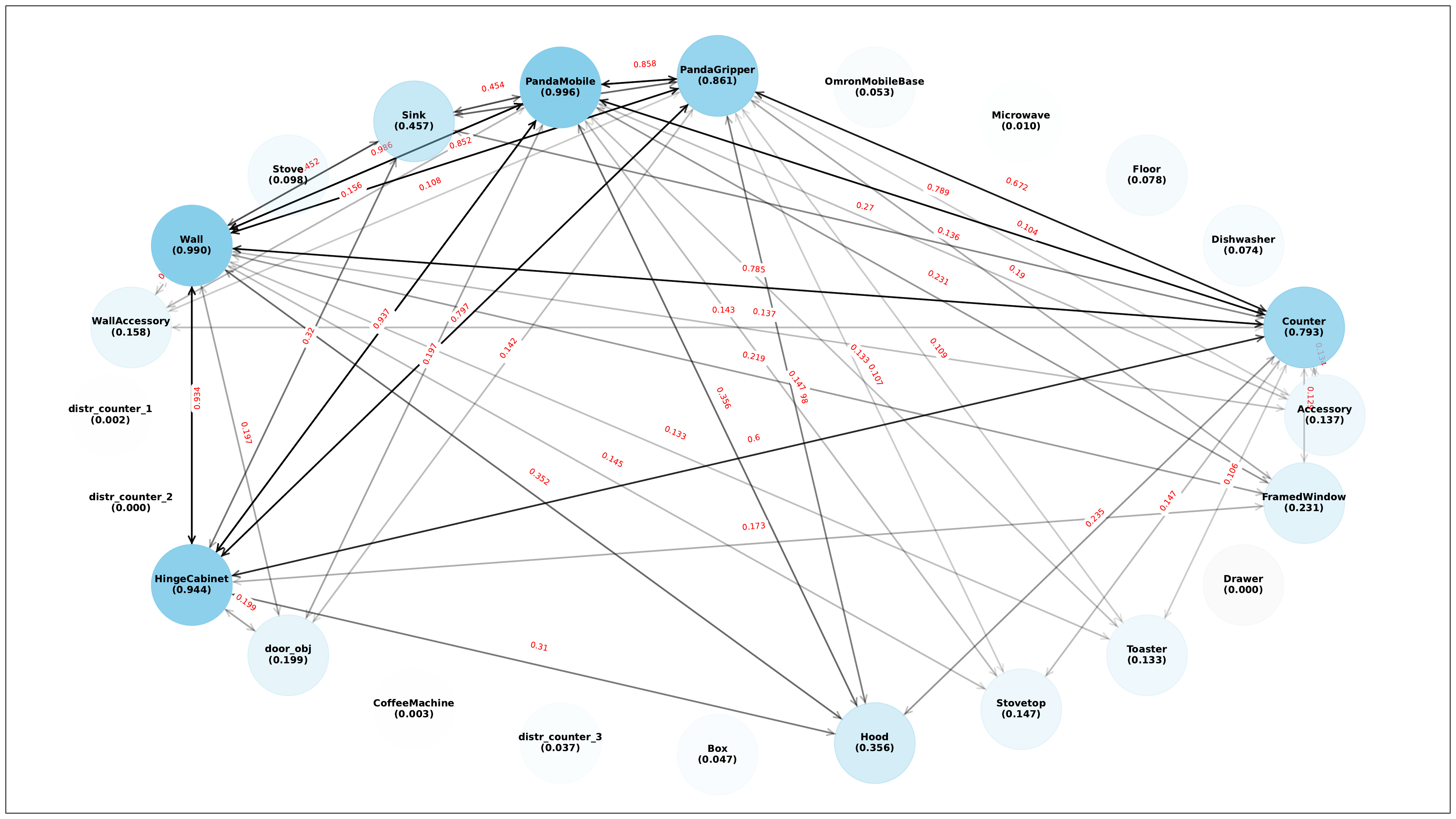}
        \caption{Task-explanation graph for model \textit{TopK Node Removal}, task \textit{OpenDoubleDoor}.}
        \label{fig:ex_double_door}
    \end{subfigure}
    \hfill
    \begin{subfigure}[t]{0.45\textwidth}
        \centering
        \includegraphics[width=\textwidth]{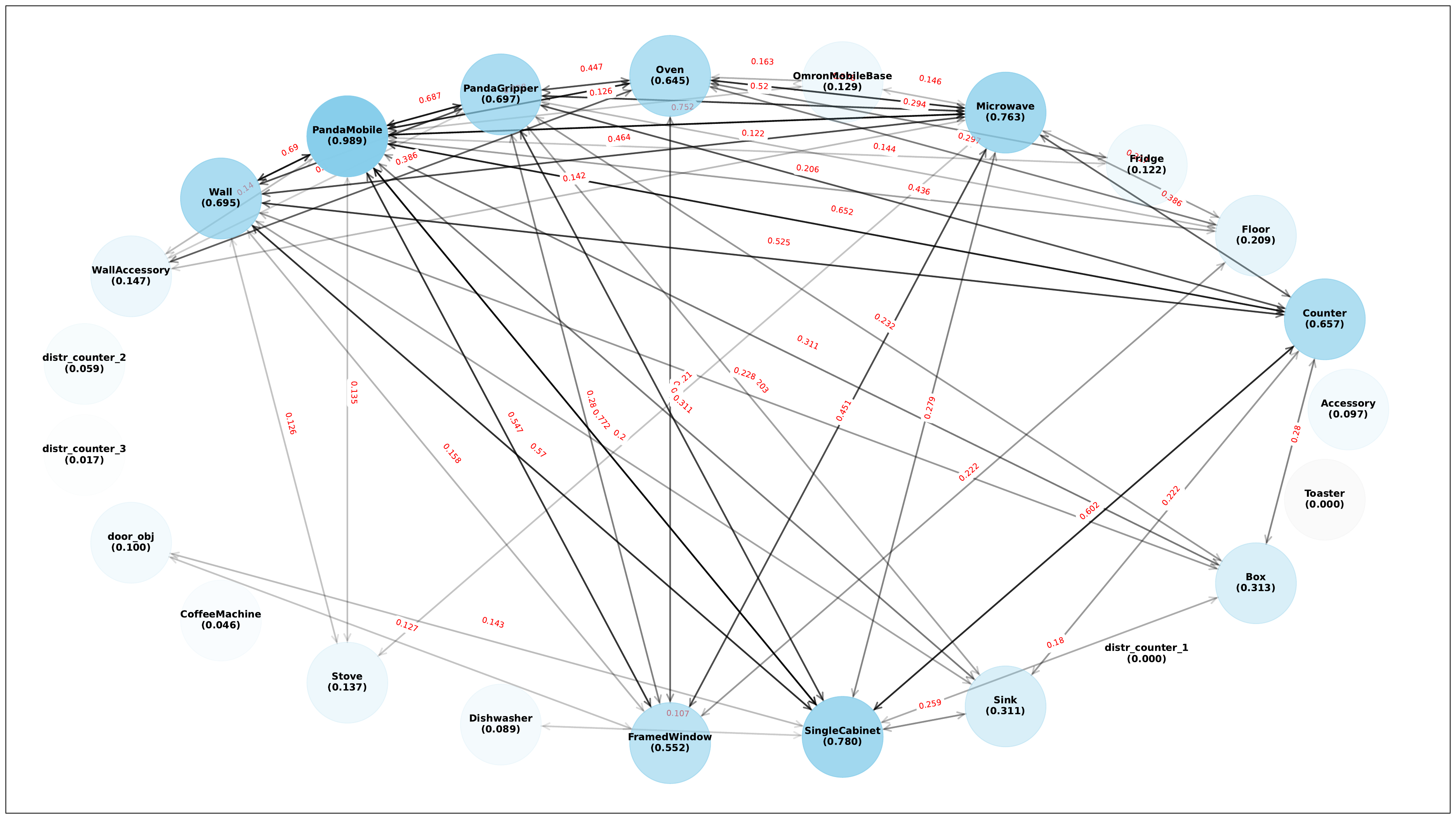}
        \caption{Task-explanation graph for model \textit{TopK Node Removal}, task \textit{CloseSingleDoor}.}
        \label{fig:ex_topk_single_door}
    \end{subfigure}
    
    \vspace{0.5em}
    
    \begin{subfigure}[t]{0.45\textwidth}
        \centering
        \includegraphics[width=\textwidth]{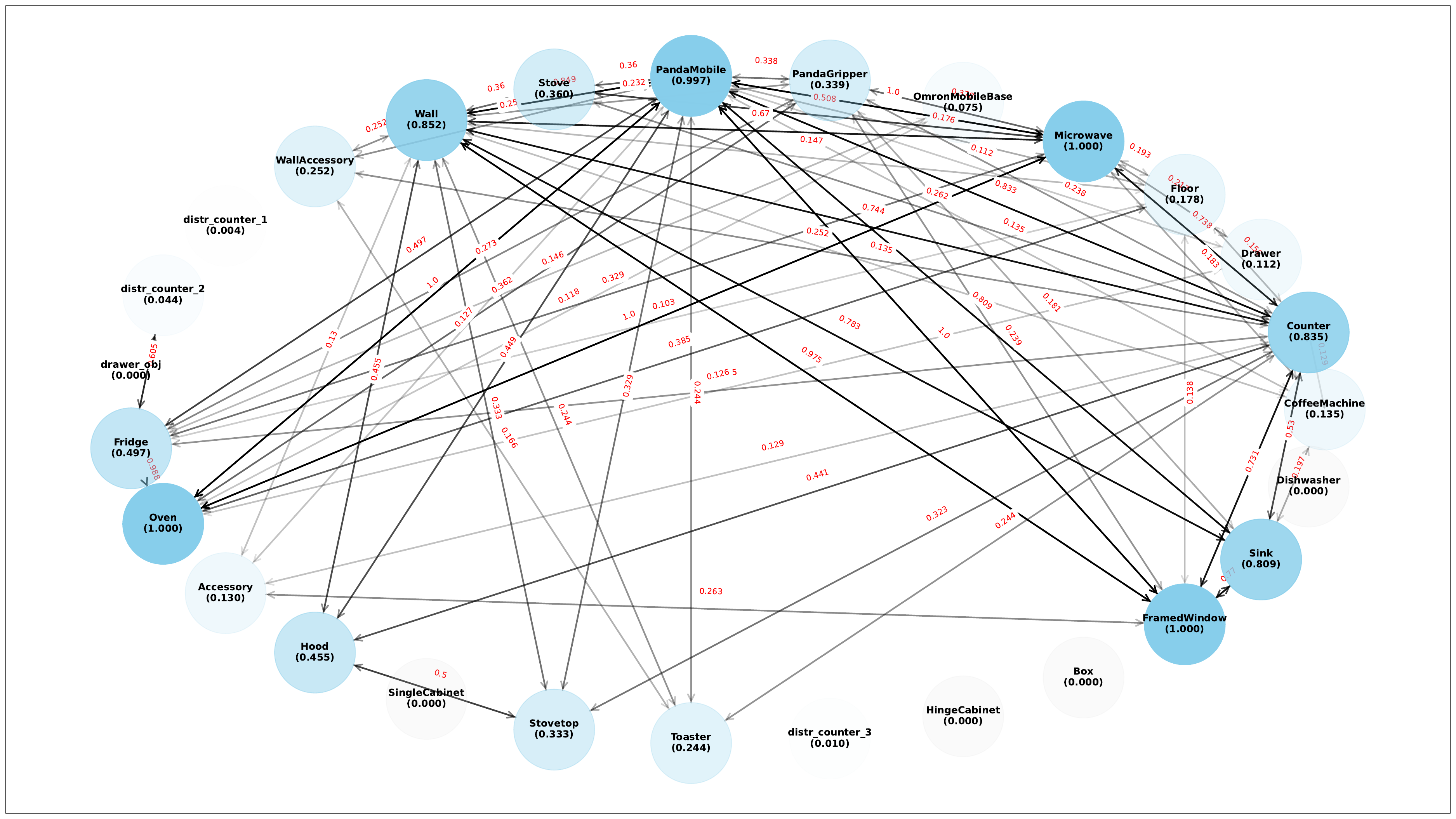}
        \caption{Task-explanation graph for model \textit{TopK Node Removal}, task \textit{CloseDrawer}.}
        \label{fig:ex_close_drawer}
    \end{subfigure}
    \hfill
    \begin{subfigure}[t]{0.45\textwidth}
        \centering
        \includegraphics[width=\textwidth]{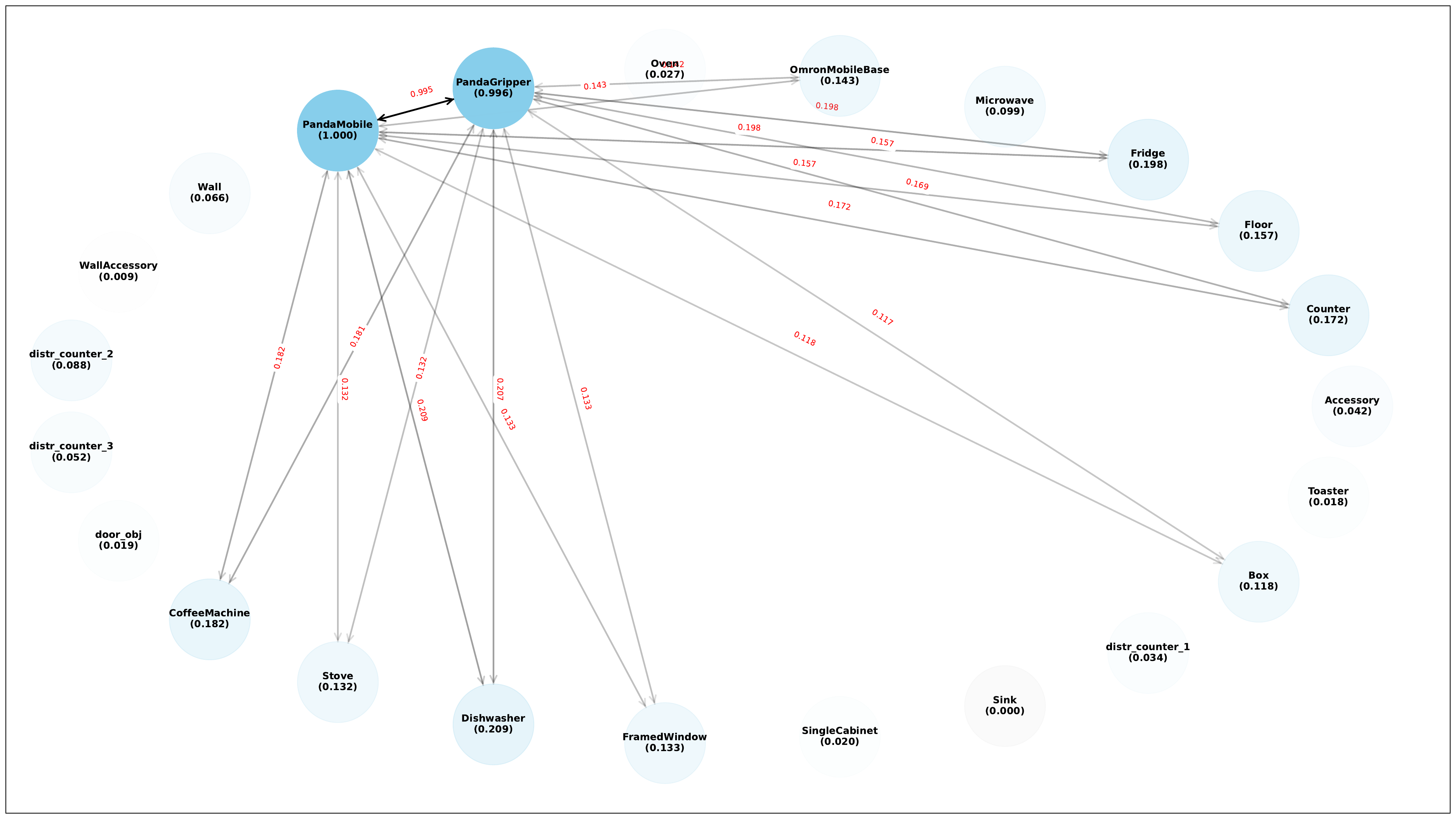}
        \caption{Task-explanation graph for model \textit{Instruction-Grounded Node Removal}, task \textit{CloseSingleDoor}.}
        \label{fig:ex_ig_single_door}
    \end{subfigure}

    \caption{Comparing \cref{fig:ex_double_door} (a hard task) and \cref{fig:ex_topk_single_door} (an easier task) shows that the explanation for the harder task is more consistent, suggesting the model learns to identify more informative nodes as task complexity increases. These learned sub-graphs are effective at revealing significant dataset biases. For example, in \cref{fig:ex_close_drawer}, the model includes many unimportant nodes, which indicates that it may be exploiting spurious correlations in the dataset for that task. Similarly, in \cref{fig:ex_ig_single_door}, the instruction-grounded model only includes the robot nodes and omits the target objects, which strongly suggests a positional bias in the data.}
    \label{fig:four_task_expl_graphs}
\end{figure*}

\subsection{Explainable Model Behaviour}
\label{sec:explainabiliy}
Learning the sub-graph as an intermediate representation during training enables an analysis of the model's understanding of the current observed scene.
We differentiate three main sub-graph types, displayed in \Cref{fig:subgraphs}: (1) Human expected sub-graph, (2) sub-graph with distractor nodes and (3) sub-graph with missing nodes.
Our learnable approaches do not consistently produce the human expected sub-graph, because the model itself learns the graph, which can result in deviations. 
These deviations, which fall into categories (2) and (3), are the primary source of insight, as they allow us to analyse the model's actual decision-making process.
\Cref{fig:four_task_expl_graphs} displays four task-explanation sub-graphs generated using \Cref{eq:ExplanationGraphGeneration}.
Further results are displayed in \Cref{app:xai_statistics}.
The graphs are based on the 100 rollouts of the specific model on the given task.
For the \textit{CloseSingleDoor} task either the door of the cabinet or the microwave has to be closed.
In case of the \textit{OpenDoubleDoor}, always two cabinet doors have to be opened.
We mainly consider explanation graphs from the TopK approach, as this does not introduce prior biases showcasing the models intent.

\paragraph{Sub-Graph with Distractor Nodes} As seen in \Cref{fig:four_task_expl_graphs}, the learned sub-graphs often include distractor objects (all sub-figures except \cref{fig:ex_ig_single_door}).
For instance, \cref{fig:ex_double_door} displays the sub-graph for the \textit{OpenDoubleDoor} task, one of the poor-performing task settings. 
In this case, the overall sub-graph is relatively consistent but includes objects like \textit{Wall} and \textit{Counter}, which are not directly related to opening doors.
In \Cref{fig:ex_close_drawer}, the sub-graph for the \textit{CloseDrawer} task is less consistent, yet the model's performance is high (81\% success rate). 
This model consistently includes unrelated objects, such as the \textit{Oven}, \textit{FramedWindow} and \textit{Microwave}. 
These observations lead to the assumption that the model exploits \textbf{spurious correlations} in the given training data, where seemingly unimportant objects provide enough information to solve the given task.

\paragraph{Sub-Graph with Missing Nodes} The sub-graphs for the \textit{CloseSingleDoor} task reveal a compelling insight when comparing the TopK \cref{fig:ex_topk_single_door} and \ac{sir} \cref{fig:ex_ig_single_door}.
The TopK model \cref{fig:ex_topk_single_door} includes the relevant objects, \textit{SingleCabinet} and \textit{Microwave}, with a high frequency (over 70\%), although it also selects various irrelevant nodes. 
\ac{sir} \cref{fig:ex_ig_single_door}, however, learns an entirely different sub-graph: it almost exclusively selects the \textit{PandaMobile} and \textit{PandaGripper} nodes. 
It consistently excludes the primary task objects \textit{SingleCabinet} and \textit{Microwave}.
Despite this complete omission of key objects, \ac{sir} outperforms the TopK model by over 5\%.
This result strongly indicates that the dataset contains significant \textbf{positional biases}.
The model has learned it can succeed by executing a fixed trajectory based only on its own gripper's state, rendering the actual position of the target door obsolete.
This also holds for the \textit{CloseDrawer} task in \cref{fig:ex_close_drawer}, where \textit{Drawer} is only included 11\% of the time.

\paragraph{Interpretable Behaviour} \ac{sir} provides insight into action generation models by learning an intermediate sub-graph end-to-end. 
The key insights do not come from "correct" sub-graphs, but from their deviations from a human-expected graph.
This allows us to evaluate the model's behaviour: Is it succeeding for the right reasons, or is it focusing on unimportant features?
Furthermore, when a model succeeds despite a deviating sub-graph, it reveals critical insights into the dataset and how the model has learned to exploit it. 
These observations are only possible due to \ac{sir}'s end-to-end nature.
This contrasts with methods like a VLM or \ac{llm} used to pre-filter objects \cite{compose_sgs}. 
A VLM would always include logically important objects and exclude unimportant ones, creating a "clean" sub-graph.
This, however, would completely mask the underlying dataset or model biases that \ac{sir}'s end-to-end learned sub-graphs successfully expose.
All these points can be used to answer \textbf{RQ4}, clearly demonstrating the effectiveness of our proposed method to understand model and even dataset intrinsic.

\section{Conclusion}
In this paper, we introduced \ac{sir}, a method to generate and use learned, sparsified \acp{sg} as an intermediate representation for robot policy learning in \ac{gcil}.
Our investigation shows that graph-based representations achieve higher average success rates than image-based baselines and are a highly effective architecture for integrating diverse modalities like point clouds. 
Furthermore, graph-based policies are significantly more robust to distractor objects, showing almost no performance degradation where image-based policies fail.
Our most critical finding is that the learned, sparsified sub-graphs serve as a powerful tool for model and dataset debugging.
By analysing when the model's graph deviates from human intuition, such as by including distractor nodes or excluding key task-relevant nodes, we successfully identified significant spurious correlations and positional biases in the dataset.
This demonstrates that an end-to-end learned, explainable representation like \ac{sir} can uncover flaws in training data.
Such insights would be completely masked by non-end-to-end methods, like \ac{vlm} pre-filtering, which would always select the "correct" objects and hide these biases.
For future work, we plan to extend our node selection method to allow the model to learn how many nodes are important, rather than relying on heuristics.
We further want to reduce the reliance on ground-truth data for graph generation and plan to fine-tune 2D foundation models to predict model masks.

\section{Acknowledgment}
The work was funded by the German Research Foundation
(DFG) – 448648559. 
This work is supported by the Helmholtz Association Initiative and Networking Fund under the KiKIT Pilot Program Core-Informatics.
The authors gratefully acknowledge the support of the Robotics Institute Germany (RIG).
The authors gratefully acknowledge the computing time provided on the high-performance computer HoreKa by the National High-Performance Computing Center at KIT (NHR@KIT). This center is jointly supported by the Federal Ministry of Education and Research and the Ministry of Science, Research and the Arts of Baden-Württemberg, as part of the National High-Performance Computing (NHR) joint funding program (https://www.nhr-verein.de/en/our-partners). HoreKa is partly funded by the German Research Foundation (DFG).
{
    \small
    \bibliographystyle{ieeenat_fullname}
    \bibliography{main}
}

\clearpage
\setcounter{page}{1}
\maketitlesupplementary

\section{Additional information}
Source code will be made publicly available upon acceptance.
Our \ac{gnn} implementation employs residual connections and normalization to mitigate oversmoothing, a phenomenon to which small, fully-connected graphs are particularly susceptible.

\subsection{Pre-Trained Models}
\label{app:pre_trained_models}
The pre-trained models for generating Cropped-Image-Features and Point-Cloud-Features were trained using human demonstration data from RoboCasa \cite{robocasa2024}. We extracted all objects from the scene at every 10th time step to train these vision networks using a reconstruction loss.
For Cropped-Image-Features alone, we employ a ResNet18 \cite{resnet} backbone with an embedding dimension of 256. However, when combining Cropped-Image-Features with BB-Coordinates, we use a smaller ResNet8 backbone with an embedding dimension of 37. This lower dimension was chosen specifically to match the size of the object label vectors (one-hot encoded length of 37), allowing the visual features to serve as a direct replacement for symbolic labels.

\subsection{Graph Sparsification}

Usually, graph sparsification methods are based on the \ac{ib} principle \citep{yin_recipe_2025} or an L0-regularization (or L1/L2) on the edge weights \citep{ye_sparse_2023} (in our case: node weights). However, we empirically find that such methods are not applicable in our case, as they either keep all or zero nodes in the sub-graph. Regarding our finding that the dataset is biased, we attribute this phenomenon to the lack of discriminability of the information content of the node features. I.e., since there are strong correlations between objects' positions in the scene (or the position is not relevant at all), there is no extreme advantage of choosing one node over another, which makes it difficult to select one node over another to keep in the sub-graph. This is further complicated by the necessity for hard masks, which can hinder differentiability and optimizations, but are required in interpretability-seeking settings. This demonstrates the necessity for our soft histogram loss, which encourages an explicit ranking for the nodes, and the TopK-selection mechanism to ensure that there are neither all nor zero nodes in the sub-graphs.

\section{Additional Evaluation Results}
The following sections present results for \ac{sir} using \ac{bc}, generated graphs and the CALVIN benchmark \cite{calvin}. 
Additionally, we provide fine-grained performance metrics for all 24 RoboCasa tasks and tabular results for the distractor experiments.
\Cref{tab:main_paper_all} displays the average results of all models using \ac{mdt} as the action generator.

\begin{table}[t]
\caption{Ablation results on RoboCasa using generated graphs, which are predicted using a fine-tuned DETR model. Models are not trained on these generated graphs, these are only swapped for ground truth information during rollout.}
\label{tab:generated_graphs}
\centering
\small
\begin{tabular}{@{}lcc@{}}
\toprule
\textbf{Task} & \textbf{Cropped-Img} & \textbf{BB + Crop-Img} \\
\midrule
Pick/Place (8) & 0.06 & 0.06 \\
Doors (4) & 11.75 & 18.13 \\
Drawers (2) & 34.00 & 28.75 \\
Knobs (2) & 9.00 & 12.00 \\
Levers (3) & 39.67 & 35.83 \\
Buttons (3) & 12.33 & 10.50 \\
Insert (2) & 3.25 & 3.25 \\
\midrule
\textbf{Avg (24)} & 12.33 & 12.50 \\
\bottomrule
\end{tabular}
\end{table}

\subsection{Generated Graphs Results}
\label{app:gen_results}
This approach is called \textit{generated graphs} and relies on a fine-tuned DETR for graph generation.
The DETR model is used during rollout to predict the objects in a given scene and their segmentation masks, which the bounding boxes can be derived from.
We evaluated using generated graphs on \ac{fcg} on RoboCasa using the Split-View approach.
Models were not trained on the generated graphs, but on the ground truth information.
This introduces a distribution shift, because the generated graphs will be imperfect, which is also visible in the performance drop, seen in \Cref{tab:generated_graphs}.
Compared to using ground truth graphs, the models drop over 4 percent for only using Cropped-Image-Features and 3.4 percent for using BB-Coordinates and Cropped-Image-Features.
These results could have two causes, either the model can not handle the distribution shift from ground truth to generated graphs or generated graphs in general will lead to a worse performance.
In future experiments, we also want to \textbf{train} networks on generated graphs to have a clear answer to this observation.

\begin{table*}[t]
\caption{Results for using \ac{bc}-Transformer as the action generator on RoboCasa across all 24 tasks.}
\label{tab:robocasa_bc}
\centering
\small
\resizebox{\linewidth}{!}{
\begin{tabular}{@{}lccccccccr@{}}
\toprule
Feature Input & Pick/Place (8) & Doors (4) & Drawers (2) & Knobs (2) & Levers (3) & Buttons (3) & Insert (2) & \textbf{Avg (24)} \\
\midrule
\textit{Baselines} & & & & & & & & \\
Image & \textbf{0.94} & 25.63 & 36.50 & 5.75 & 22.33 & \textbf{24.00} & \textbf{7.00} & 14.48 \\
\midrule[0.5pt]
\textit{Split-View Graph - Fully-Connected} & & & & & & & & \\
Cropped-Img & 0.13 & 12.63 & 31.50 & 11.75 & 34.50 & 6.00 & 5.75 & 11.25 \\
BB-Coordinates + Label & 0.00 & 11.88 & 23.00 & \textbf{16.25} & 15.00 & 0.00 & 0.25 & 7.33 \\
BB-Coordinates + Cropped-Img & 0.06 & 9.00 & 24.75 & 6.50 & 36.33 & 5.33 & 3.00 & 9.58 \\
\midrule[0.5pt]
\textit{Fusion Graph - Fully-Connected} & & & & & & & & \\
BB-Coordinates + Cropped-Img & 0.25 & 18.13 & 34.5 & 11.25 & 31.83 & 5.0 & 0.3 & 11.77 \\
\midrule[0.5pt]
\textit{Sparse Graph Methods} & & & & & & & & \\
TopK & 0.31 & 17.25 & 36.75 & 10.0 & 37.83 & 6.67 & 2.25 & 12.63 \\
\ac{sir} (Ours) & 0.38 & \textbf{29.63} & \textbf{42.25} & 14.5 & \textbf{43.33} & 6.83 & 2.5 & \textbf{16.27} \\
\bottomrule
\end{tabular}
}
\end{table*}

\subsection{RoboCasa - Behaviour Cloning Results}
\label{app:bc_results}
The results for the \ac{bc}-Transformer as action generation model in RoboCasa can be seen in \Cref{tab:robocasa_bc}.
Using images as observation input only decreases the average result slightly compared to using \ac{mdt} as the action generation model.
In comparison, using graphs and \ac{bc}-Transformer, performance drops heavily, for all approaches.
But using \ac{sir} still results in the highest average success rate of 16.27 percent.
These observation lead to the conclusion that graph observation can be better utilized by diffusion-based methods compared to \ac{bc}-based methods, but overall achieve a higher result compared to image-based models regardles of training objective.

\begin{table}[t]
\caption{Completion scores on CALVIN using only the static camera as observation $\text{D} \rightarrow \text{D}$ for 100 Rollouts using two seeds for each model. The maximum completion number is 5.}
\label{tab:calvin_mdt_results}
\centering
\small
\begin{tabular}{@{}lc@{}}
\toprule
\textbf{Model} & \textbf{Task Completion (Max 5)} \\
\midrule
\textit{Baseline} & \\
Image & \textbf{1.5 $\pm$ 0.2} \\
\midrule[0.5pt]
\textit{Fully-Connected Graphs} & \\
Cropped-Img & 1.3 $\pm$ 0.02 \\
BB-Coordinates + Cropped-Img & 1.2 $\pm$ 0.03 \\
\midrule[0.5pt]
\textit{Sparse Graph Methods} & \\
TopK & 1.4 $\pm$ 0.02 \\
\bottomrule
\end{tabular}
\end{table}

\subsection{CALVIN - MDT Results}
\label{app:calvin_results}
CALVIN \cite{calvin} serves as our second evaluation benchmark. 
The results, presented in \Cref{tab:calvin_mdt_results}, report the average number of tasks completed out of a possible five per rollout. 
Image-based models achieve an average of 1.5 tasks, slightly outperforming the sparse TopK graph method, which completes 1.4 tasks.
It is important to note that our experiments on the CALVIN environment are still preliminary. 
We hypothesize that further fine-tuning of the sparsification methods could yield performance gains similar to those observed in RoboCasa.

\begin{table*}[t]
\caption{Distractor objects: Success rate over 100 rollouts. Higher values indicate better performance.}
\label{tab:distractor_tasks}
\centering
\small
\resizebox{\linewidth}{!}{
\begin{tabular}{@{}lcccccccc@{}}
\toprule
Feature Input & Pick/Place (8) & Doors (4) & Drawers (2) & Knobs (2) & Levers (3) & Buttons (3) & Insert (2) & \textbf{Avg (24)} \\
\midrule
\textit{Baseline} & & & & & & & & \\
Image                     & \textbf{1.19} & 25.13 & \textbf{49.75} & 7.25  & 23.67 & 17.00 & \textbf{4.75} & 14.81 \\
Image w/ Distractors      & 0.56 & 18.25 & 46.25 & 3.25  & 17.17 & 14.50 & 2.50 & 11.52 \\
\midrule[0.5pt]
\textit{Fully Connected Graph} & & & & & & & & \\
FC-Graph                & 0.06 & 28.62 & 39.25 & 14.00 & 40.00 & 18.83 & \textbf{4.75} & 16.98 \\
FC-Graph w/ Distractors & 0.38 & 21.12 & 40.00 & 17.75 & 41.50 & 18.67 & 3.75 & 16.29 \\
\midrule[0.5pt]
\textit{Sparse Graph Methods} & & & & & & & & \\
Threshold & 0.00 & 28.38 & 47.25 & 11.00 & 39.50 & 19.17 & 3.00 & 17.17  \\
Threshold w/ Distractors & 0.40 & 22.50 & 44.50 & 12.75 & 40.33 & 22.17 & 2.50 & 16.67 \\
TopK                 & 0.12 & \textbf{31.37} & 45.00 & 19.50 & 41.17 & 18.17 & 4.50 & 18.44 \\
TopK w/ Distractors  & 0.38 & 22.75 & 44.75 & 14.25 & 36.50 & 14.50 & 4.00 & 15.54 \\
\ac{sir}                  & 0.12 & 30.25 & 46.25 & 16.50 & 48.50 & \textbf{21.83} & \textbf{4.75} & \textbf{19.50} \\
\ac{sir} w/ Distractors   & 0.56 & 23.12 & 47.50 & \textbf{21.75} & \textbf{51.83} & 20.67 & 4.25 & 19.23 \\
\bottomrule
\end{tabular}
}
\end{table*}

\subsection{Distractor Objects}
The introduction of distractor objects in RoboCasa leads to a clear decline in performance for both the image baseline and the TopK sparsification approach, displayed in \Cref{tab:distractor_tasks}. 
While success rates for the \textit{Pick and Place} task appear to increase for all models except the image baseline, the absolute scores are too low to be considered conclusive. 
In the \textit{Insert} task, performance decreases across the board, though graph-based models exhibit significantly smaller degradation.

\begin{table*}[t]
\caption{The results represent success rate over 100 rollouts with 2 models trained on different seeds for Pick and Place tasks.}
\label{tab:pnp_tasks}
\centering
\small
\resizebox{\linewidth}{!}{
\begin{tabular}{@{}lccccccccc@{}}
\toprule
\shortstack[l]{Feature \\ Input} & 
\shortstack{PnPCab \\ ToCounter} & 
\shortstack{PnPCounter \\ ToCab} & 
\shortstack{PnPMicrowave \\ ToCounter} & 
\shortstack{PnPCounter \\ ToMicrowave} & 
\shortstack{PnPSink \\ ToCounter} & 
\shortstack{PnPCounter \\ ToSink} & 
\shortstack{PnPStove \\ ToCounter} & 
\shortstack{PnPCounter \\ ToStove} & 
\textbf{Average} \\
\midrule
\textit{Baselines} & & & & & & & & & \\
Image & 3.0 & 1.5 & 2.5 & 0.0 & 0.0 & 0.5 & 0.5 & 1.5 & 1.19 \\
Image + FiLM & 1.0 & 0.0 & 0.0 & 0.5 & 0.0 & 0.25 & 0.5 & 0.5 & 0.34 \\
Own Pretrained Image & 0.5 & 0.0 & 0.0 & 0.0 & 0.0 & 0.5 & 0.0 & 0.0 & 0.13 \\
Point Clouds & 0.0 & 0.0 & 0.0 & 0.0 & 0.0 & 0.0 & 0.0 & 0.0 & 0.0 \\
Image + Point Clouds & 1.5 & 0.0 & 0.0 & 0.5 & 0.5 & 1.0 & 0.5 & 0.5 & 0.56 \\
\midrule[0.5pt]
\textit{Split-View Graph - Fully-Connected} & & & & & & & & & \\
Cropped-Img & 1.0 & 0.0 & 0.5 & 0.0 & 0.0 & 0.0 & 0.5 & 0.0 & 0.25 \\
BB-Coordinates + Label & 0.5 & 0.0 & 0.0 & 0.0 & 0.0 & 0.5 & 0.0 & 0.0 & 0.13 \\
BB-Coordinates + Cropped-Img & 1.5 & 0.0 & 0.0 & 0.0 & 0.5 & 0.0 & 0.0 & 0.0 & 0.25 \\
Point Clouds & 0.5 & 0.0 & 0.0 & 0.0 & 0.0 & 0.0 & 0.0 & 0.0 & 0.06 \\
Label + Point Clouds & 1.0 & 0.5 & 0.0 & 0.0 & 0.5 & 0.67 & 0.5 & 0.0 & 0.40 \\
Cropped-Img + Point Clouds & 1.0 & 0.0 & 0.0 & 0.0 & 0.0 & 1.0 & 0.0 & 0.0 & 0.25 \\
\midrule[0.5pt]
\textit{Fusion Graph - Fully-Connected} & & & & & & & & & \\
BB-Coordinates + Label & 0.0 & 0.0 & 0.0 & 0.0 & 0.0 & 0.5 & 0.0 & 0.0 & 0.06 \\
BB-Coordinates + Cropped-Img & 0.0 & 0.0 & 0.0 & 0.0 & 0.5 & 0.0 & 0.0 & 0.0 & 0.06 \\
\midrule[0.5pt]
\textit{Sparsification Methods} & & & & & & & & & \\
Random Node Removal & 0.0 & 0.0 & 0.0 & 0.0 & 0.0 & 0.0 & 0.0 & 0.0 & 0.0 \\
Naive NR (no soft histogram loss) & 0.0 & 0.0 & 0.0 & 0.0 & 0.0 & 0.0 & 0.0 & 0.0 & 0.0 \\
Threshold & 0.0 & 0.0 & 0.0 & 0.0 & 0.0 & 0.0 & 0.0 & 0.0 & 0.0 \\
TopK & 0.5 & 0.0 & 0.0 & 0.0 & 0.0 & 0.5 & 0.0 & 0.0 & 0.12 \\
SIR (Ours) & 0.5 & 0.0 & 0.0 & 0.0 & 0.5 & 0.0 & 0.0 & 0.0 & 0.12 \\
\bottomrule
\end{tabular}
}
\end{table*}

\begin{table*}[t]
\caption{Door tasks: success rate over 100 rollouts with 2 models trained on different seeds.}
\label{tab:door_tasks}
\centering
\small
\begin{tabular}{@{}lccccc@{}}
\toprule
\shortstack[l]{Feature \\ Input} & 
\shortstack{Close \\ Single Door} & 
\shortstack{Open \\ Single Door} & 
\shortstack{Close \\ Double Door} & 
\shortstack{Open \\ Double Door} & 
\textbf{Average} \\
\midrule
\textit{Baselines} & & & & & \\
Image & 63.0 & 6.0 & 30.0 & 1.5 & 25.13 \\
Image + FiLM & 64.5 & 7.5 & 36.0 & 3.0 & 27.75 \\
Own Pretrained Image & 55.5 & 4.0 & 8.0 & 0.5 & 17.00 \\
Point Clouds & 0.0 & 0.0 & 0.0 & 0.0 & 0.00 \\
Image + Point Clouds & 59.0 & 6.0 & 33.5 & 2.5 & 25.25 \\
\midrule[0.5pt]
\textit{Split-View Graph - Fully-Connected} & & & & & \\
Cropped-Img & 48.5 & 16.0 & 17.5 & 1.5 & 20.88 \\
BB-Coordinates + Label & 54.0 & 3.5 & 18.5 & 0.5 & 19.13 \\
BB-Coordinates + Cropped-Img & 59.0 & 14.5 & 18.0 & 1.0 & 23.13 \\
Point Clouds & 21.5 & 0.0 & 24.0 & 0.0 & 11.38 \\
Label + Point Clouds & 24.5 & 6.5 & 31.5 & 1.0 & 15.88 \\
Cropped-Img + Point Clouds & 39.5 & 10.0 & 26.0 & 2.0 & 19.38 \\
\midrule[0.5pt]
\textit{Fusion Graph - Fully-Connected} & & & & & \\
BB-Coordinates + Label & 47.5 & 3.0 & 6.0 & 0.0 & 14.12 \\
BB-Coordinates + Cropped-Img & 64.0 & 18.5 & 29.0 & 3.0 & 28.62 \\
\midrule[0.5pt]
\textit{Sparsification Methods} & & & & & \\
Random Node Removal & 9.0 & 7.0 & 6.5 & 0.0 & 5.62 \\
Naive NR (no soft histogram loss) & 33.5 & 0.5 & 0.0 & 0.0 & 8.50 \\
Threshold & 61.5 & 17.5 & 32.0 & 2.5 & 28.38 \\
TopK & 60.0 & 25.0 & 35.0 & 5.5 & 31.37 \\
SIR (Ours) & 65.5 & 22.0 & 33.0 & 0.5 & 30.25 \\
\bottomrule
\end{tabular}
\end{table*}

\begin{table*}[t]
\caption{Drawer tasks: success rate over 100 rollouts with 2 models trained on different seeds.}
\label{tab:drawer_tasks}
\centering
\small
\begin{tabular}{@{}lccc@{}}
\toprule
\shortstack[l]{Feature \\ Input} & 
\shortstack{Close \\ Drawer} & 
\shortstack{Open \\ Drawer} & 
\textbf{Average} \\
\midrule
\textit{Baselines} & & & \\
Image & 85.5 & 14.0 & 49.75 \\
Image + FiLM & 80.5 & 10.0 & 45.25 \\
Own Pretrained Image & 68.5 & 10.0 & 39.25 \\
Point Clouds & 2.0 & 0.0 & 1.0 \\
Image + Point Clouds & 82.0 & 7.5 & 44.75 \\
\midrule[0.5pt]
\textit{Split-View Graph - Fully-Connected} & & & \\
Cropped-Img & 67.5 & 7.0 & 37.25 \\
BB-Coordinates + Label & 61.0 & 2.0 & 31.5 \\
BB-Coordinates + Cropped-Img & 63.0 & 7.5 & 35.25 \\
Point Clouds & 51.0 & 7.5 & 29.25 \\
Label + Point Clouds & 57.5 & 7.0 & 32.25 \\
Cropped-Img + Point Clouds & 70.0 & 11.5 & 40.75 \\
\midrule[0.5pt]
\textit{Fusion Graph - Fully-Connected} & & & \\
BB-Coordinates + Label & 55.0 & 8.0 & 31.5 \\
BB-Coordinates + Cropped-Img & 75.0 & 3.5 & 39.25 \\
\midrule[0.5pt]
\textit{Sparsification Methods} & & & \\
Random Node Removal & 18.5 & 0.5 & 9.5 \\
Naive NR (no soft histogram loss) & 46.0 & 1.0 & 23.5 \\
Threshold & 87.5 & 7.0 & 47.25 \\
TopK & 81.5 & 8.5 & 45.0 \\
SIR (Ours) & 80.5 & 12.0 & 46.25 \\
\bottomrule
\end{tabular}
\end{table*}

\begin{table*}[t]
\caption{Stove tasks: success rate over 100 rollouts with 2 models trained on different seeds.}
\label{tab:stove_tasks}
\centering
\small
\begin{tabular}{@{}lccc@{}}
\toprule
\shortstack[l]{Feature \\ Input} & 
\shortstack{TurnOn \\ Stove} & 
\shortstack{TurnOff \\ Stove} & 
\textbf{Average} \\
\midrule
\textit{Baselines} & & & \\
Image & 12.5 & 2.0 & 7.25 \\
Image + FiLM & 17.0 & 5.0 & 11.00 \\
Own Pretrained Image & 9.5 & 5.0 & 7.25 \\
Point Clouds & 1.0 & 1.5 & 1.25 \\
Image + Point Clouds & 9.0 & 2.0 & 5.50 \\
\midrule[0.5pt]
\textit{Split-View Graph - Fully-Connected} & & & \\
Cropped-Img & 25.0 & 5.0 & 15.00 \\
BB-Coordinates + Label & 21.5 & 5.5 & 13.50 \\
BB-Coordinates + Cropped-Img & 21.5 & 7.0 & 14.25 \\
Point Clouds & 28.0 & 9.0 & 18.50 \\
Label + Point Clouds & 23.5 & 6.5 & 15.00 \\
Cropped-Img + Point Clouds & 12.0 & 5.0 & 8.50 \\
\midrule[0.5pt]
\textit{Fusion Graph - Fully-Connected} & & & \\
BB-Coordinates + Label & 17.0 & 9.0 & 13.00 \\
BB-Coordinates + Cropped-Img & 22.5 & 5.5 & 14.00 \\
\midrule[0.5pt]
\textit{Sparsification Methods} & & & \\
Random Node Removal & 8.0 & 2.5 & 5.25 \\
Naive NR (no soft histogram loss) & 8.5 & 5.0 & 6.75 \\
Threshold & 15.5 & 6.5 & 11.0 \\
TopK & 26.0 & 13.0 & 19.50 \\
SIR (Ours) & 26.5 & 6.5 & 16.50 \\
\bottomrule
\end{tabular}
\end{table*}

\begin{table*}[t]
\caption{Sink tasks: success rate over 100 rollouts with 2 models trained on different seeds.}
\label{tab:sink_tasks}
\centering
\small
\begin{tabular}{@{}lcccc@{}}
\toprule
\shortstack[l]{Feature \\ Input} & 
\shortstack{TurnOn \\ Sink Faucet} & 
\shortstack{TurnOff \\ Sink Faucet} & 
\shortstack{Turn \\ Sink Spout} & 
\textbf{Average} \\
\midrule
\textit{Baselines} & & & & \\
Image & 22.0 & 24.0 & 25.0 & 23.67 \\
Image + FiLM & 32.0 & 32.5 & 34.0 & 32.83 \\
Own Pretrained Image & 11.5 & 25.5 & 19.5 & 18.83 \\
Point Clouds & 27.5 & 2.5 & 29.5 & 19.83 \\
Image + Point Clouds & 21.5 & 18.5 & 22.0 & 20.67 \\
\midrule[0.5pt]
\textit{Split-View Graph - Fully-Connected} & & & & \\
Cropped-Img & 36.0 & 64.5 & 44.5 & 48.33 \\
BB-Coordinates + Label & 12.0 & 17.5 & 34.0 & 21.17 \\
BB-Coordinates + Cropped-Img & 30.5 & 51.5 & 39.0 & 40.33 \\
Point Clouds & 15.5 & 24.0 & 51.5 & 30.33 \\
Label + Point Clouds & 26.0 & 31.5 & 53.0 & 36.83 \\
Cropped-Img + Point Clouds & 24.5 & 55.0 & 43.0 & 40.83 \\
\midrule[0.5pt]
\textit{Fusion Graph - Fully-Connected} & & & & \\
BB-Coordinates + Label & 11.5 & 25.5 & 44.5 & 27.17 \\
BB-Coordinates + Cropped-Img & 23.0 & 54.5 & 42.5 & 40.00 \\
\midrule[0.5pt]
\textit{Sparsification Methods} & & & & \\
Random Node Removal & 5.0 & 21.5 & 24.5 & 17.00 \\
Naive NR (no soft histogram loss) & 15.0 & 33.0 & 39.5 & 29.17 \\
Threshold & 20.5 & 54.5 & 43.5 & 39.5 \\
TopK & 22.0 & 55.0 & 46.5 & 41.17 \\
SIR (Ours) & 31.5 & 69.0 & 45.0 & 48.50 \\
\bottomrule
\end{tabular}
\end{table*}

\begin{table*}[t]
\caption{Button tasks: success rate over 100 rollouts with 2 models trained on different seeds.}
\label{tab:button_tasks}
\centering
\small
\begin{tabular}{@{}lcccc@{}}
\toprule
\shortstack[l]{Feature \\ Input} & 
\shortstack{TurnOff \\ Microwave} & 
\shortstack{TurnOn \\ Microwave} & 
\shortstack{Coffee \\ PressButton} & 
\textbf{Average} \\
\midrule
\textit{Baselines} & & & & \\
Image & 22.0 & 15.0 & 14.0 & 17.00 \\
Image + FiLM & 27.5 & 14.5 & 8.5 & 16.83 \\
Own Pretrained Image & 9.5 & 9.5 & 11.5 & 10.17 \\
Point Clouds & 24.0 & 5.5 & 3.0 & 10.83 \\
Image + Point Clouds & 17.5 & 15.0 & 13.0 & 15.17 \\
\midrule[0.5pt]
\textit{Split-View Graph - Fully-Connected} & & & & \\
Cropped-Img & 22.5 & 13.5 & 15.5 & 17.17 \\
BB-Coordinates + Label & 12.0 & 8.5 & 2.5 & 7.67 \\
BB-Coordinates + Cropped-Img & 25.5 & 14.5 & 13.0 & 17.67 \\
Point Clouds & 12.0 & 13.5 & 3.5 & 9.67 \\
Label + Point Clouds & 11.5 & 10.5 & 3.0 & 8.33 \\
Cropped-Img + Point Clouds & 15.0 & 15.0 & 18.0 & 16.00 \\
\midrule[0.5pt]
\textit{Fusion Graph - Fully-Connected} & & & & \\
BB-Coordinates + Label & 12.0 & 15.5 & 5.5 & 11.00 \\
BB-Coordinates + Cropped-Img & 30.5 & 17.5 & 8.5 & 18.83 \\
\midrule[0.5pt]
\textit{Sparsification Methods} & & & & \\
Random Node Removal & 13.5 & 7.5 & 4.0 & 8.33 \\
Naive NR (no soft histogram loss) & 21.5 & 9.5 & 10.5 & 13.83 \\
Threshold & 31.5 & 15.0 & 11.0 & 19.17 \\
TopK & 25.0 & 18.5 & 11.0 & 18.17 \\
SIR (Ours) & 38.5 & 14.0 & 13.0 & 21.83 \\
\bottomrule
\end{tabular}
\end{table*}

\begin{table*}[t]
\caption{Coffee tasks: success rate over 100 rollouts with 2 models trained on different seeds.}
\label{tab:coffee_tasks}
\centering
\small
\begin{tabular}{@{}lccc@{}}
\toprule
\shortstack[l]{Feature \\ Input} & 
\shortstack{Coffee \\ Serve Mug} & 
\shortstack{Coffee \\ Setup Mug} & 
\textbf{Average} \\
\midrule
\textit{Baselines} & & & \\
Image & 9.5 & 0.0 & 4.75 \\
Image + FiLM & 3.0 & 0.0 & 1.50 \\
Own Pretrained Image & 4.0 & 0.0 & 2.00 \\
Point Clouds & 2.5 & 0.0 & 1.25 \\
Image + Point Clouds & 4.0 & 0.5 & 2.25 \\
\midrule[0.5pt]
\textit{Split-View Graph - Fully-Connected} & & & \\
Cropped-Img & 12.0 & 1.0 & 6.50 \\
BB-Coordinates + Label & 9.0 & 0.5 & 4.75 \\
BB-Coordinates + Cropped-Img & 12.5 & 1.5 & 7.00 \\
Point Clouds & 4.5 & 0.0 & 2.25 \\
Label + Point Clouds & 3.0 & 0.5 & 1.75 \\
Cropped-Img + Point Clouds & 11.0 & 1.5 & 6.25 \\
\midrule[0.5pt]
\textit{Fusion Graph - Fully-Connected} & & & \\
BB-Coordinates + Label & 4.5 & 0.5 & 2.50 \\
BB-Coordinates + Cropped-Img & 8.5 & 1.0 & 4.75 \\
\midrule[0.5pt]
\textit{Sparsification Methods} & & & \\
Random Node Removal & 3.5 & 0.0 & 1.75 \\
Naive NR (no soft histogram loss) & 7.0 & 0.0 & 3.50 \\
Threshold & 6.0 & 0.0 & 3.0 \\
TopK & 8.5 & 0.5 & 4.50 \\
SIR (Ours) & 9.0 & 0.5 & 4.75 \\
\bottomrule
\end{tabular}
\end{table*}

\subsection{Per Task Results}
\label{app:per_task_results}
High average reward for the grouped tasks from the RoboCasa \cite{robocasa2024} paper does not mean that the model performs well in each subtask.
Therefore, we included all single task results in the following tables: \textit{Pick and Place} in \Cref{tab:pnp_tasks}, \textit{Doors} in \Cref{tab:door_tasks}, \textit{Drawers} in \Cref{tab:drawer_tasks}, \textit{Knobs} in \Cref{tab:stove_tasks}, \textit{Levers} in \Cref{tab:sink_tasks}, \textit{Buttons} in \Cref{tab:button_tasks} and \textit{Insert} in \Cref{tab:coffee_tasks}.
Each task grouping (except \textit{Pick and Place}) includes tasks which are easier solvable and harder tasks.
\ac{sir} only performs best on 4 single atomic tasks, which indicates that the higher average performance is distributed across all single tasks.

\begin{table*}[t]
\caption{Success rate on RoboCasa across all 24 tasks over 100 rollouts with 2 models trained on different seeds.}
\label{tab:main_paper_all}
\centering
\small
\resizebox{\linewidth}{!}{
\begin{tabular}{@{}lccccccccr@{}}
\toprule
Feature Input & Pick/Place (8) & Doors (4) & Drawers (2) & Knobs (2) & Levers (3) & Buttons (3) & Insert (2) & \textbf{Avg (24)} \\
\midrule
\textit{Baselines} & & & & & & & & \\
Image & \textbf{1.19} & 25.13 & \textbf{49.75} & 7.25 & 23.67 & 17.00 & 4.75 & 14.81 \\
Image + FiLM & 0.34 & 27.75 & 45.25 & 11.00 & 32.83 & 16.83 & 1.50 & 15.85 \\
Own Pretrained Image & 0.13 & 17.00 & 39.25 & 7.25 & 18.83 & 10.17 & 2.00 & 10.11 \\
Point Clouds & 0.00 & 0.00 & 1.00 & 1.25 & 19.83 & 10.83 & 1.25 & 4.13 \\
Image + Point Clouds & 0.56 & 25.25 & 44.75 & 5.50 & 20.67 & 15.17 & 2.25 & 13.25 \\
\midrule[0.5pt]
\textit{Split-View Graph - Fully-Connected} & & & & & & & & \\
Cropped-Img & 0.25 & 20.88 & 37.25 & 15.00 & 48.33 & 17.17 & 6.50 & 16.65 \\
BB-Coordinates + Label & 0.13 & 19.13 & 31.50 & 13.50 & 21.17 & 7.67 & 4.75 & 10.98 \\
BB-Coordinates + Cropped-Img & 0.25 & 23.13 & 35.25 & 14.25 & 40.33 & 17.67 & \textbf{7.00} & 15.90 \\
Point Clouds & 0.06 & 11.38 & 29.25 & 18.50 & 30.33 & 9.67 & 2.25 & 11.08 \\
Label + Point Clouds & 0.40 & 15.88 & 32.25 & 15.00 & 36.83 & 8.33 & 1.75 & 12.50 \\
Cropped-Img + Point Clouds & 0.25 & 19.38 & 40.75 & 8.50 & 40.83 & 16.00 & 6.25 & 15.04 \\
\midrule[0.5pt]
\textit{Fusion Graph - Fully-Connected} & & & & & & & & \\
BB-Coordinates + Label & 0.06 & 14.12 & 31.50 & 13.00 & 27.17 & 11.00 & 2.50 & 11.06 \\
BB-Coordinates + Cropped-Img & 0.06 & 28.62 & 39.25 & 14.00 & 40.00 & 18.83 & 4.75 & 16.98 \\
\midrule[0.5pt]
\textit{Sparsification Methods} & & & & & & & & \\
Random Node Removal & 0.00 & 5.62 & 9.50 & 5.25 & 17.00 & 8.33 & 1.75 & 5.48 \\
Naive NR (no soft histogram loss) & 0.00 & 8.50 & 23.50 & 6.75 & 29.17 & 13.83 & 3.50 & 9.60 \\
Threshold & 0.0 & 28.38 & 47.25 & 11.0 & 39.5 & 19.17 & 3.0 & 17.17 \\
TopK & 0.12 & \textbf{31.37} & 45.00 & \textbf{19.50} & 41.17 & 18.17 & 4.50 & 18.44 \\
SIR (Ours) & 0.12 & 30.25 & 46.25 & 16.50 & \textbf{48.50} & \textbf{21.83} & 4.75 & \textbf{19.50} \\
\bottomrule
\end{tabular}
}
\end{table*}

\begin{table*}[t]
\caption{MDT Policy Success Rate ($\%$) for all 24 Atomic Tasks (100 Rollouts) using the In-hand camera as additional modality.}
\label{tab:atomic_task_comparison_inhand}
\centering
\footnotesize
\setlength{\tabcolsep}{3pt}
\begin{tabular}{@{}lccc@{}}
\toprule
\textbf{Atomic Task} & \textbf{Image + In-hand} & \textbf{FC-Graph + In-hand} & \textbf{\ac{sir} + In-hand} \\
\midrule
\multicolumn{4}{@{}l}{\textbf{Pick and Place (8)}} \\
PnPCab $\rightarrow$ Ctr & 8.5 & 6.5 & 5.5 \\
PnP Ctr $\rightarrow$ Cab & 11.5 & 9.0 & 6.0 \\
PnPMW $\rightarrow$ Ctr & 4.0 & 2.0 & 0.5 \\
PnP Ctr $\rightarrow$ MW & 12.0 & 5.0 & 1.0 \\
PnPSink $\rightarrow$ Ctr & 6.5 & 5.0 & 7.5 \\
PnP Ctr $\rightarrow$ Sink & 6.0 & 9.5 & 8.0 \\
PnP Stove $\rightarrow$ Ctr & 1.5 & 8.0 & 6.5 \\
PnP Ctr $\rightarrow$ Stove & 1.0 & 2.0 & 1.0 \\
\midrule
\multicolumn{4}{@{}l}{\textbf{Doors (4)}} \\
Close Single Door & 67.5 & 75.0 & 76.0\\
Open Single Door & 29.5 & 38.5 & 49.0 \\
Close DoubleDoor & 39.0 & 33.5 & 18.0 \\
Open DoubleDoor & 15.5 & 10.5 & 2.0 \\
\midrule
\multicolumn{4}{@{}l}{\textbf{Drawers (2)}} \\
Close Drawer & 93.5 & 84.0 & 89.5 \\
Open Drawer & 25.5 & 25.5 & 25.0 \\
\midrule
\multicolumn{4}{@{}l}{\textbf{Knobs (Stove) (2)}} \\
TurnOn Stove & 9.0 & 35.5 & 31.5 \\
TurnOff Stove & 4.5 & 14.5 & 12.0 \\
\midrule
\multicolumn{4}{@{}l}{\textbf{Levers (Sink) (3)}} \\
TurnOn Sink Faucet & 34.5 & 41.0 & 36.0 \\
TurnOff Sink Faucet & 25.0 & 36.0 & 40.0 \\
Turn Sink Spout & 31.0 & 26.0 & 29.5 \\
\midrule
\multicolumn{4}{@{}l}{\textbf{Buttons (3)}} \\
Turn Off Microwave & 47.0 & 55.5 & 58.5 \\
Turn On Microwave & 40.0 & 52.0 & 60.5 \\
Coffee Press Button & 49.0 & 62.0 & 67.5 \\
\midrule
\multicolumn{4}{@{}l}{\textbf{Insertion (Coffee) (2)}} \\
Coffee Serve Mug & 22.5 & 29.0 & 27.5 \\
Coffee Setup Mug & 4.0 & 3.5 & 4.0 \\
\midrule
Average & 24.5 & 30.5 & 27.6 \\
\bottomrule
\end{tabular}
\end{table*}

\section{Additional Explainability Results}
We further present additional explainability results for \ac{sir} on 6 tasks in RoboCasa, not present in the main paper, as well as a Grad-CAM visualization of the image baseline models.

\subsection{Sub-Graph Statistics}
\label{app:xai_statistics}
The generated explanations by the different models are presented in \cref{tab:FiveMostRelevantNodes_PnP}, \cref{tab:FiveMostRelevantNodes_Door}, \cref{tab:FiveMostRelevantNodes_Drawer}, \cref{tab:FiveMostRelevantNodes_Knobs}, \cref{tab:FiveMostRelevantNodes_Lever}, \cref{tab:FiveMostRelevantNodes_Button}, \cref{tab:FiveMostRelevantNodes_Insertion}.
Each column represents one of the three possibilities for the network to extract the desired sub-graph.
The results show that the four examples in \cref{fig:four_task_expl_graphs} are not cherry-picked, but that over all tasks a specific trend can be observed.

\subsection{Sub-Graph Visualizations}
\begin{figure*}[t]
    \centering
    \includegraphics[width=0.9\textwidth]{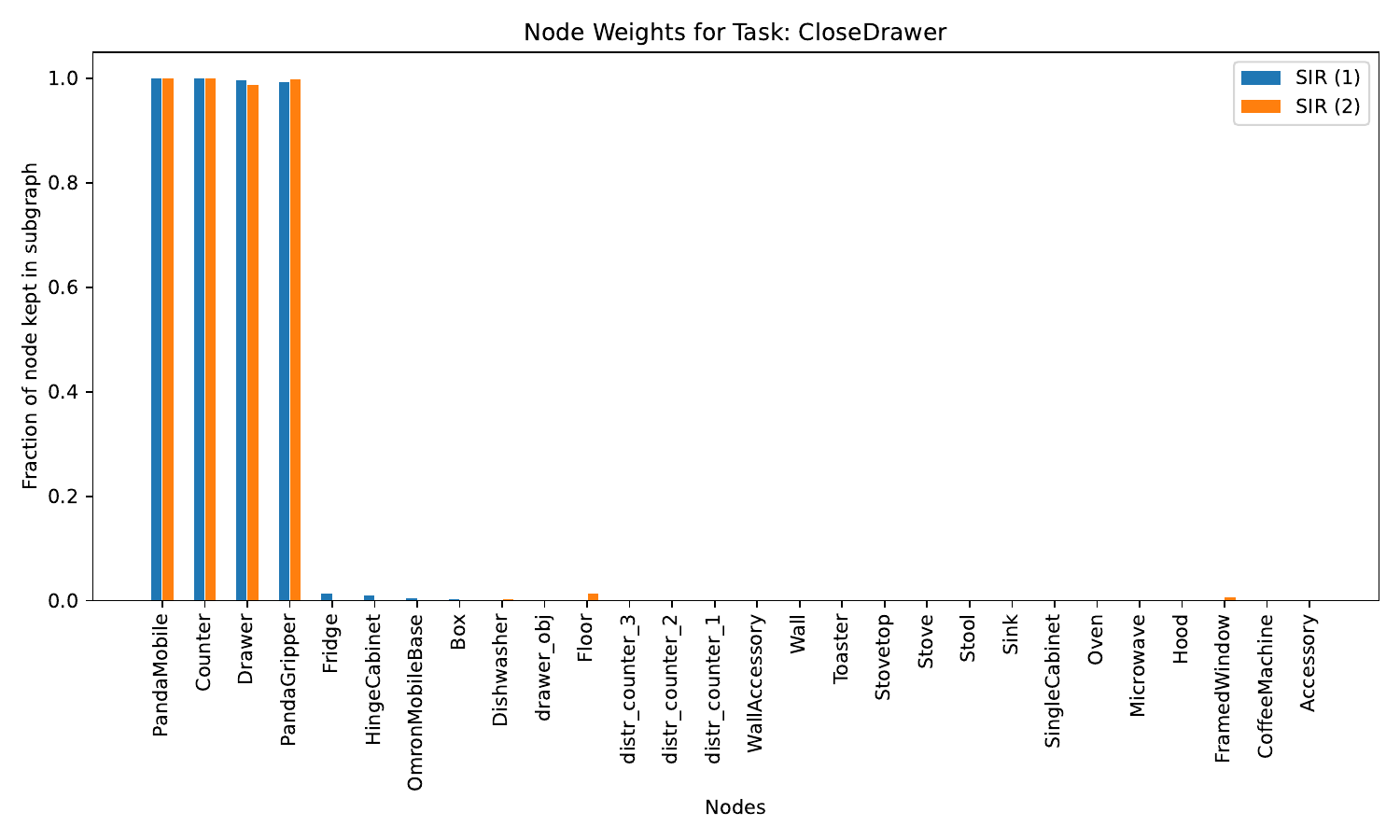}
    \caption{Percentage of nodes to be kept in the sub-graph per seeded model \textit{CloseDrawer}.
    In contrast to \Cref{fig:ex_close_drawer}, \ac{sir} is able to include the drawer in the graph. This stands in no conflict with our previous claim that the drawer is not important, as the instruction-grounded method is explicitly \textit{encouraged} by the additional loss to include the drawer. This further stands in no conflict with our results, where even in the instruction-grounded approach, the relevant nodes are not included, as the instruction-grounded approach is not \textit{forced} to use the important nodes.  }
    \label{fig:CloseDrawerKeptNodePercentages}
\end{figure*}

\begin{figure*}[t]
    \centering
    \includegraphics[width=0.9\textwidth]{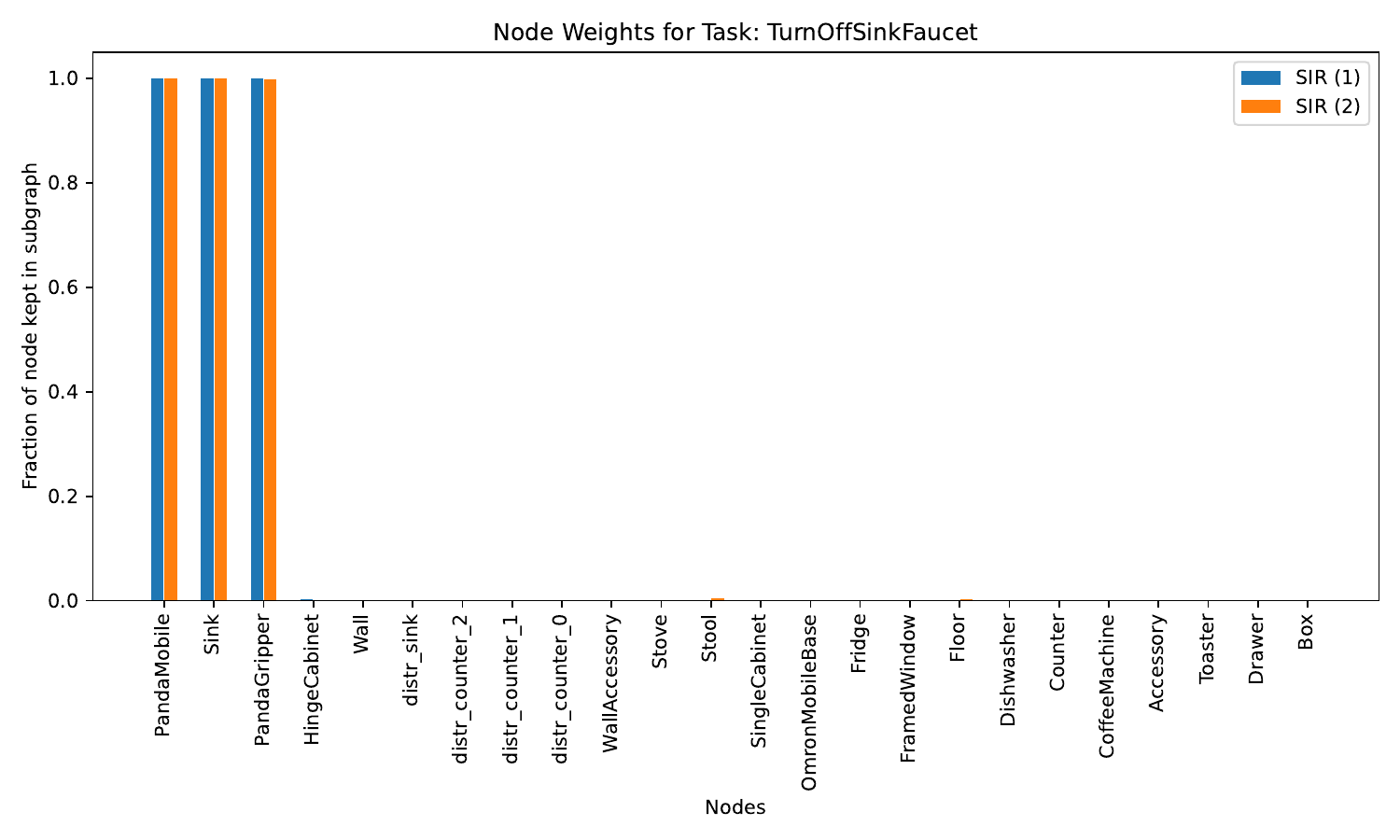}
    \caption{Percentage of nodes to be kept in the sub-graph per seeded model \textit{TurnOffSinkFaucet}. }
    \label{fig:TurnOffSinkFaucetKeptNodePercentages}
\end{figure*}

\begin{figure*}[t]
    \centering
    \includegraphics[width=0.9\textwidth]{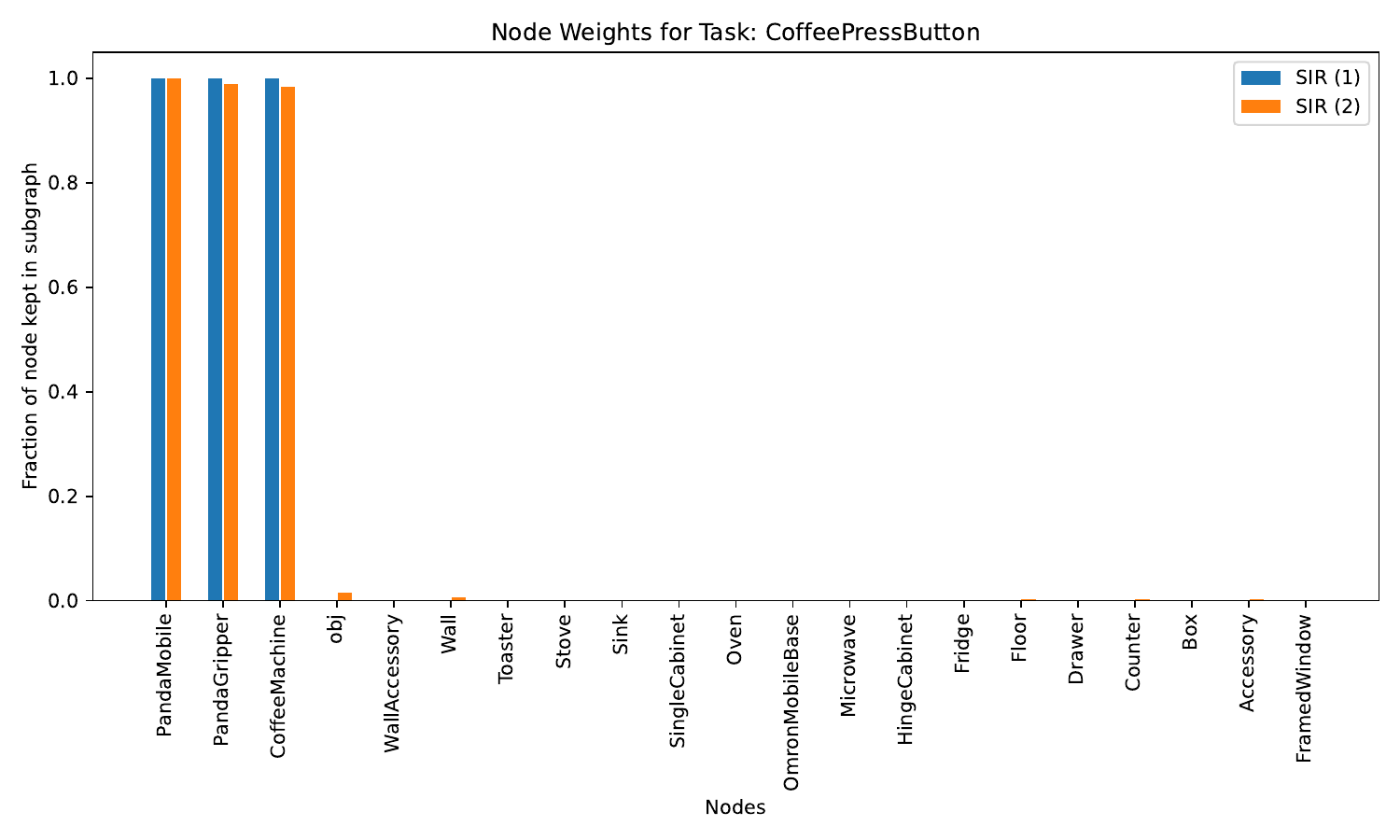}
    \caption{Percentage of nodes to be kept in the sub-graph per seeded model \textit{CoffeePressButton}.}
    \label{fig:CoffeePressButtonKeptNodePercentages}
\end{figure*}

\begin{figure*}[t]
    \centering
    \includegraphics[width=0.9\textwidth]{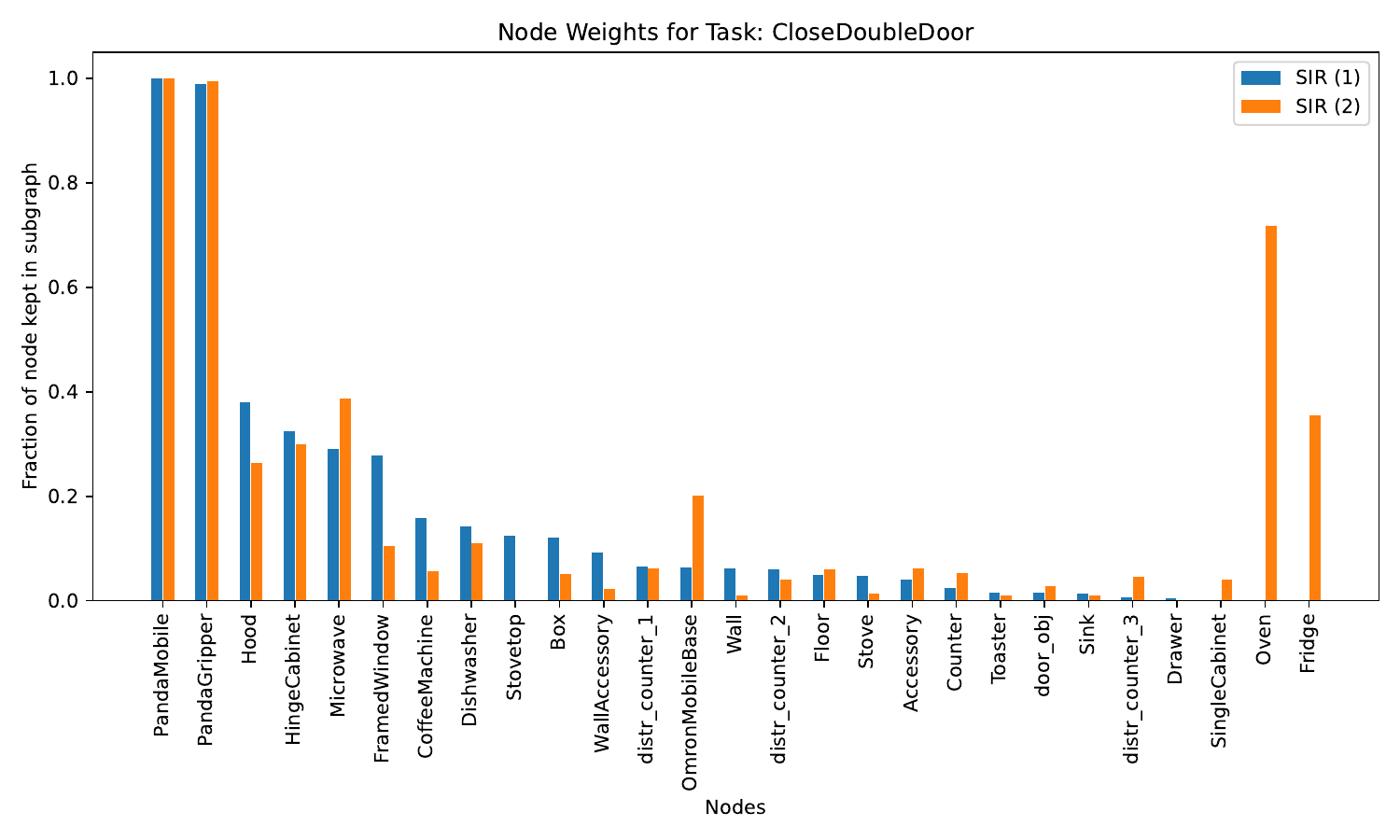}
    \caption{Percentage of nodes to be kept in the sub-graph per seeded model \textit{CloseDoubleDoor}.
    Similar to \Cref{fig:ex_ig_single_door}, here also the object of which the door has to be close (HingeCabinet) is not always selected to be in the graph.}
    \label{fig:CloseDoubleDoorKeptNodePercentages}
\end{figure*}

\begin{figure*}[t]
    \centering
    \includegraphics[width=0.9\textwidth]{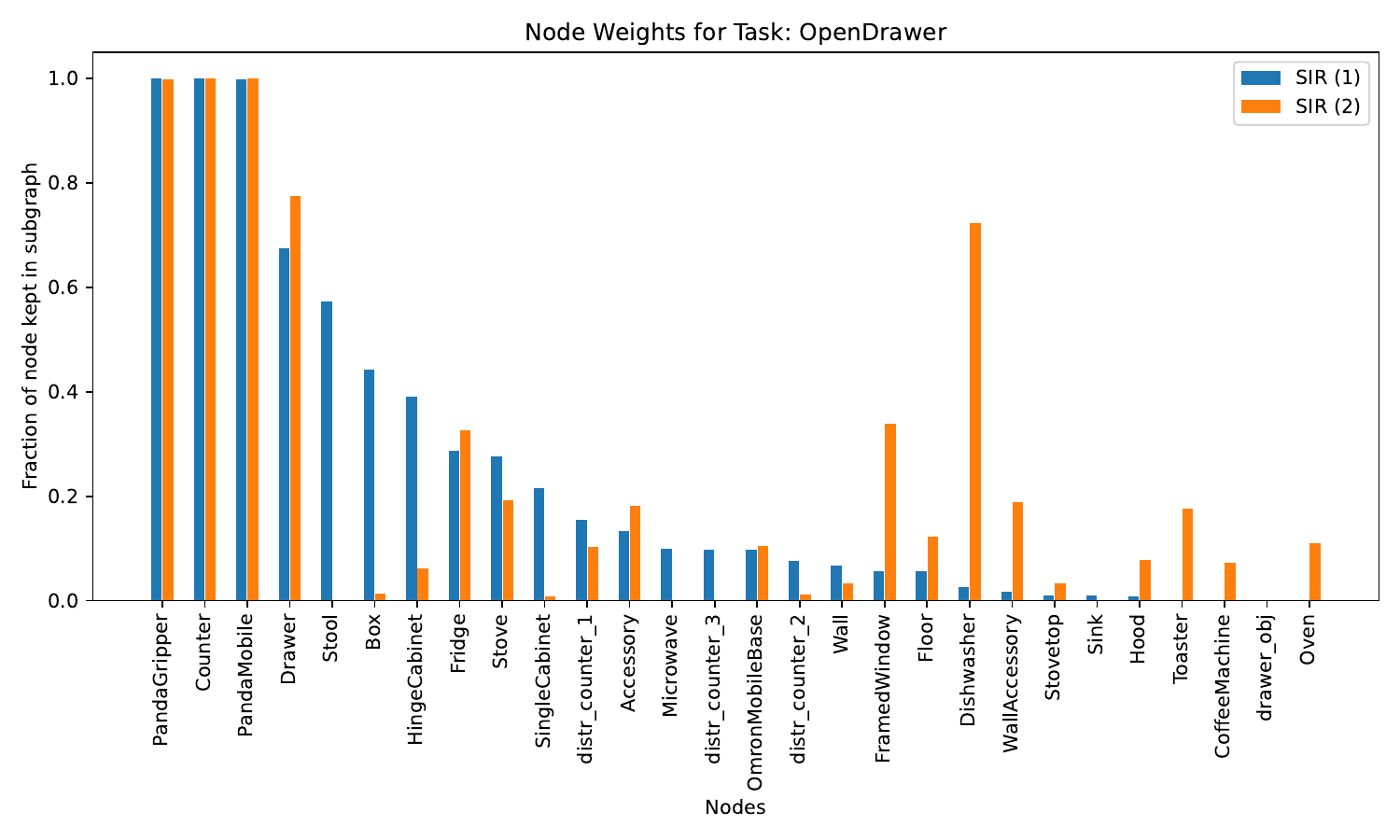}
    \caption{Percentage of nodes to be kept in the sub-graph per seeded model \textit{OpenDrawer}.
    Although the important nodes are included, other nodes are not consistently removed. }
    \label{fig:OpenDrawerKeptNodePercentages}
\end{figure*}

\begin{figure*}[t]
    \centering
    \includegraphics[width=0.9\textwidth]{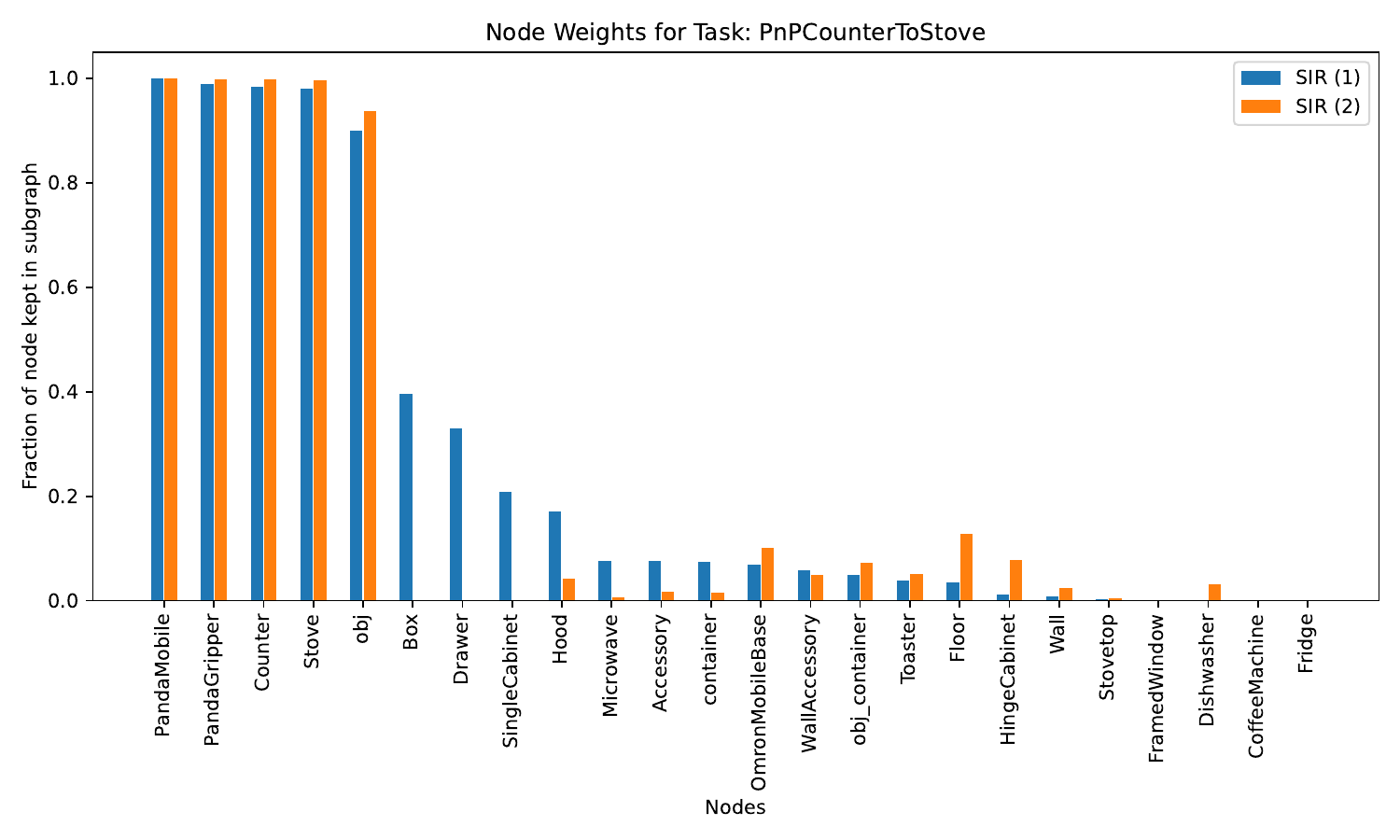}
    \caption{Percentage of nodes to be kept in the sub-graph per seeded model in \textit{PnPCounterToStove}. }
    \label{fig:PnPCounterToStoveKeptNodePercentages}
\end{figure*}

The observed explanations fall into two distinct categories based on the model's adherence to pre-defined important objects. 
The first group (\cref{fig:CloseDrawerKeptNodePercentages,fig:TurnOffSinkFaucetKeptNodePercentages,fig:CoffeePressButtonKeptNodePercentages}) demonstrates consistent reliance on the designated important nodes. 
In contrast, the second group (\cref{fig:CloseDoubleDoorKeptNodePercentages,fig:OpenDrawerKeptNodePercentages,fig:PnPCounterToStoveKeptNodePercentages}) exhibits significant variance across training seeds. While \textit{PandaGripper} and \textit{PandaMobile} remain constant, other object selections appear arbitrary.
This suggests that in tasks where the model diverges from the expected nodes yet maintains high performance, it is exploiting underlying dataset biases to solve the task.
Crucially, this inference is valid only for tasks where the model outperforms baselines, as arbitrary focus in low-performing models likely indicates a failure to learn the task.

\begin{table*}[t]
\caption{Five most relevant nodes per task and model – PnP.}
\label{tab:FiveMostRelevantNodes_PnP}
\centering
\resizebox{\linewidth}{!}{
\begin{tabular}{l|l|l|l}
Task & TopK & SIR & Threshold \\
\hline
PnPCabToCounter & \shortstack{PandaMobile (0.999) \\ HingeCabinet (1.000) / Hood (0.985) \\ Wall (0.947) / PandaGripper (0.893) \\ Counter (0.918) / Toaster (0.763) \\ SingleCabinet (0.844) / distr\_counter (0.689) } & \shortstack{SingleCabinet (1.000) / HingeCabinet (1.000) \\ PandaMobile (1.000) \\ HingeCabinet (1.000) / Counter (0.986) \\ Counter (1.000) / PandaGripper (0.977) \\ PandaGripper (0.985) / SingleCabinet (0.936) } & \shortstack{PandaMobile (1.000) \\ PandaGripper (0.964) / FramedWindow (0.950) \\ SingleCabinet (0.909) / PandaGripper (0.912) \\ Toaster (0.891) / Sink (0.690) \\ FramedWindow (0.503) / Toaster (0.672) } \\
\hline
PnPCounterToCab & \shortstack{PandaMobile (0.997) \\ HingeCabinet (1.000) / Toaster (0.952) \\ Wall (0.992) / obj (0.829) \\ SingleCabinet (0.907) / PandaGripper (0.822) \\ Counter (0.836) / distr\_counter (0.747) } & \shortstack{HingeCabinet (1.000) / Counter (1.000) \\ PandaMobile (1.000) \\ Counter (1.000) / PandaGripper (0.996) \\ PandaGripper (0.995) / HingeCabinet (0.917) \\ SingleCabinet (0.939) } & \shortstack{PandaMobile (0.999) \\ PandaGripper (0.927) \\ SingleCabinet (0.924) / Toaster (0.672) \\ Toaster (0.882) / FramedWindow (0.575) \\ FramedWindow (0.870) / SingleCabinet (0.569) } \\
\hline
PnPCounterToMicrowave & \shortstack{Microwave (0.999) / PandaMobile (0.988) \\ PandaMobile (0.999) / obj\_container (0.892) \\ Oven (0.849) / obj (0.795) \\ container (0.791) / distr\_counter (0.768) \\ Fridge (0.714) / PandaGripper (0.734) } & \shortstack{PandaMobile (1.000) \\ Microwave (0.999) \\ PandaGripper (0.951) / Counter (0.982) \\ Counter (0.951) / PandaGripper (0.980) \\ obj (0.660) } & \shortstack{Stovetop (1.000) / PandaMobile (0.998) \\ PandaMobile (0.983) / PandaGripper (0.749) \\ PandaGripper (0.829) / obj\_container (0.679) \\ obj\_container (0.800) / FramedWindow (0.568) \\ container (0.649) / Oven (0.526) } \\
\hline
PnPCounterToSink & \shortstack{PandaMobile (0.999) \\ Sink (0.974) / obj (0.815) \\ CoffeeMachine (0.937) / distr\_counter (0.803) \\ Wall (0.915) / PandaGripper (0.709) \\ Counter (0.874) / distr\_sink (0.704) } & \shortstack{PandaMobile (0.999) \\ Counter (0.998) \\ Sink (0.995) \\ PandaGripper (0.953) \\ obj (0.843) } & \shortstack{PandaMobile (1.000) \\ PandaGripper (0.832) / CoffeeMachine (0.912) \\ FramedWindow (0.726) / PandaGripper (0.823) \\ distr\_counter (0.601) / WallAccessory (0.564) \\ WallAccessory (0.509) / FramedWindow (0.461) } \\
\hline
PnPCounterToStove & \shortstack{Microwave (1.000) / PandaMobile (0.982) \\ PandaMobile (0.997) / obj\_container (0.961) \\ Wall (0.954) / obj (0.836) \\ Drawer (0.943) / PandaGripper (0.819) \\ Stovetop (0.939) / WallAccessory (0.444) } & \shortstack{PandaMobile (1.000) \\ PandaGripper (0.989) / Counter (0.999) \\ Counter (0.984) / PandaGripper (0.999) \\ Stove (0.989) \\ obj (0.919) } & \shortstack{PandaMobile (0.991) \\ Fridge (0.957) / obj\_container (0.892) \\ PandaGripper (0.883) \\ obj\_container (0.823) / CoffeeMachine (0.752) \\ WallAccessory (0.656) } \\
\hline
PnPMicrowaveToCounter & \shortstack{PandaMobile (0.988) \\ Microwave (1.000) / container (0.839) \\ Oven (0.865) / obj (0.776) \\ Fridge (0.805) / PandaGripper (0.769) \\ Wall (0.799) / distr\_counter (0.606) } & \shortstack{PandaMobile (0.993) / Microwave (0.998) \\ Microwave (0.976) / Counter (0.964) \\ PandaGripper (0.945) / PandaMobile (0.961) \\ Counter (0.940) / PandaGripper (0.951) \\ obj (0.822) } & \shortstack{Sink (1.000) / PandaMobile (0.988) \\ PandaMobile (0.997) / PandaGripper (0.733) \\ PandaGripper (0.887) / Oven (0.713) \\ container (0.652) \\ Oven (0.720) / obj (0.394) } \\
\hline
PnPSinkToCounter & \shortstack{PandaMobile (0.994) \\ Sink (0.997) / container (0.916) \\ Wall (0.955) / PandaGripper (0.877) \\ FramedWindow (0.842) / distr\_counter (0.736) \\ Counter (0.792) / obj (0.717) } & \shortstack{obj (1.000) / PandaMobile (1.000) \\ Sink (0.999) / Counter (0.981) \\ PandaMobile (0.999) / Sink (0.970) \\ Counter (0.991) / obj (0.947) \\ PandaGripper (0.952) } & \shortstack{PandaMobile (0.999) / Microwave (1.000) \\ FramedWindow (0.949) / PandaMobile (1.000) \\ PandaGripper (0.812) \\ WallAccessory (0.685) / container (0.681) \\ container (0.640) / FramedWindow (0.666) } \\
\hline
PnPStoveToCounter & \shortstack{Microwave (1.000) / PandaMobile (0.991) \\ PandaMobile (1.000) / container (0.984) \\ Stovetop (0.888) / PandaGripper (0.880) \\ Wall (0.881) / Dishwasher (0.621) \\ Counter (0.638) / HingeCabinet (0.507) } & \shortstack{PandaMobile (1.000) \\ Stove (1.000) / PandaGripper (0.999) \\ Counter (1.000) / Stove (0.998) \\ PandaGripper (0.999) / Counter (0.990) \\ obj (0.958) } & \shortstack{PandaMobile (0.994) \\ PandaGripper (0.845) / container (0.849) \\ container (0.843) / PandaGripper (0.805) \\ Stove (0.761) / WallAccessory (0.543) \\ CoffeeMachine (0.667) / Stove (0.474) } \\
\hline
\end{tabular}
}
\end{table*}

\begin{table*}[t]
\caption{Five most relevant nodes per task and model – Door.}
\label{tab:FiveMostRelevantNodes_Door}
\centering
\resizebox{\linewidth}{!}{
\begin{tabular}{l|l|l|l}
Task & TopK & SIR & Threshold \\
\hline
OpenSingleDoor & \shortstack{PandaMobile (0.979) / Microwave (0.989) \\ Wall (0.855) / FramedWindow (0.972) \\ Microwave (0.802) / PandaMobile (0.841) \\ PandaGripper (0.759) / Oven (0.835) \\ Counter (0.672) / PandaGripper (0.831) } & \shortstack{PandaMobile (1.000) \\ PandaGripper (0.985) \\ CoffeeMachine (0.302) / HingeCabinet (0.324) \\ Toaster (0.249) / Microwave (0.303) \\ OmronMobileBase (0.215) / Box (0.267) } & \shortstack{PandaMobile (0.989) \\ FramedWindow (0.974) \\ PandaGripper (0.844) \\ distr\_counter\_3 (0.732) / Microwave (0.750) \\ Oven (0.631) / distr\_counter\_2 (0.638) } \\
\hline
OpenDoubleDoor & \shortstack{PandaMobile (0.996) / PandaGripper (0.904) \\ Wall (0.990) / HingeCabinet (0.901) \\ HingeCabinet (0.944) / PandaMobile (0.891) \\ PandaGripper (0.861) / Sink (0.851) \\ Counter (0.793) / Drawer (0.751) } & \shortstack{PandaMobile (1.000) \\ PandaGripper (0.989) \\ HingeCabinet (0.403) / Drawer (0.437) \\ SingleCabinet (0.299) / Oven (0.395) \\ FramedWindow (0.289) / HingeCabinet (0.377) } & \shortstack{PandaMobile (0.997) \\ HingeCabinet (0.918) / PandaGripper (0.941) \\ PandaGripper (0.881) / HingeCabinet (0.769) \\ Toaster (0.660) \\ WallAccessory (0.469) / Sink (0.705) } \\
\hline
CloseSingleDoor & \shortstack{PandaMobile (0.989) / Microwave (0.885) \\ SingleCabinet (0.780) / PandaMobile (0.872) \\ Microwave (0.763) / Oven (0.828) \\ PandaGripper (0.697) / SingleCabinet (0.775) \\ Wall (0.695) / FramedWindow (0.750) } & \shortstack{PandaMobile (1.000) \\ PandaGripper (0.997) \\ Dishwasher (0.209) / HingeCabinet (0.431) \\ Fridge (0.198) / Box (0.351) \\ CoffeeMachine (0.182) / Fridge (0.314) } & \shortstack{PandaMobile (0.998) / Stovetop (1.000) \\ FramedWindow (0.959) / PandaMobile (0.984) \\ PandaGripper (0.879) / FramedWindow (0.907) \\ SingleCabinet (0.805) / PandaGripper (0.888) \\ Oven (0.648) / SingleCabinet (0.747) } \\
\hline
CloseDoubleDoor & \shortstack{PandaMobile (0.994) / HingeCabinet (0.964) \\ Wall (0.950) / PandaMobile (0.894) \\ HingeCabinet (0.950) / PandaGripper (0.798) \\ Counter (0.798) / SingleCabinet (0.657) \\ PandaGripper (0.571) / FramedWindow (0.637) } & \shortstack{PandaMobile (1.000) \\ PandaGripper (0.991) \\ Hood (0.381) / Oven (0.718) \\ HingeCabinet (0.324) / Microwave (0.388) \\ Microwave (0.291) / Fridge (0.355) } & \shortstack{PandaMobile (0.997) \\ HingeCabinet (0.913) / PandaGripper (0.852) \\ PandaGripper (0.835) / Toaster (0.751) \\ FramedWindow (0.747) / HingeCabinet (0.744) \\ Toaster (0.571) / Sink (0.736) } \\
\hline
\end{tabular}
}
\end{table*}

\begin{table*}[t]
\caption{Five most relevant nodes per task and model – Drawer.}
\label{tab:FiveMostRelevantNodes_Drawer}
\centering
\resizebox{\linewidth}{!}{
\begin{tabular}{l|l|l|l}
Task & TopK & SIR & Threshold \\
\hline
OpenDrawer & \shortstack{Oven (1.000) / PandaMobile (0.939) \\ Microwave (1.000) / distr\_counter\_2 (0.900) \\ FramedWindow (1.000) / distr\_counter\_1 (0.873) \\ PandaMobile (0.970) / Toaster (0.782) \\ Wall (0.879) / Stovetop (0.753) } & \shortstack{PandaGripper (1.000) / PandaMobile (1.000) \\ Counter (1.000) \\ PandaMobile (0.998) / PandaGripper (0.999) \\ Drawer (0.726) \\ Stool (0.574) / Dishwasher (0.722) } & \shortstack{Oven (1.000) / PandaMobile (0.989) \\ PandaMobile (0.991) / Sink (0.885) \\ FramedWindow (0.781) \\ Sink (0.938) / OmronMobileBase (0.491) \\ Toaster (0.681) / PandaGripper (0.484) } \\
\hline
CloseDrawer & \shortstack{Oven (1.000) / distr\_counter\_2 (0.887) \\ Microwave (1.000) / PandaMobile (0.842) \\ FramedWindow (1.000) / PandaGripper (0.781) \\ PandaMobile (0.997) / Toaster (0.718) \\ Wall (0.852) / distr\_counter\_1 (0.712) } & \shortstack{PandaMobile (1.000) \\ Counter (1.000) \\ Drawer (0.996) / PandaGripper (0.999) \\ PandaGripper (0.994) / Drawer (0.987) \\ Fridge (0.014) / Floor (0.013) } & \shortstack{PandaMobile (0.980) \\ FramedWindow (0.811) / Sink (0.789) \\ Sink (0.797) / PandaGripper (0.590) \\ Toaster (0.737) / FramedWindow (0.500) \\ Oven (0.700) / OmronMobileBase (0.467) } \\
\hline
\end{tabular}
}
\end{table*}

\begin{table*}[t]
\caption{Five most relevant nodes per task and model – Knobs.}
\label{tab:FiveMostRelevantNodes_Knobs}
\centering
\resizebox{\linewidth}{!}{
\begin{tabular}{l|l|l|l}
Task & TopK & SIR & Threshold \\
\hline
TurnOnStove & \shortstack{Microwave (1.000) / Sink (1.000) \\ Stove (1.000) \\ PandaMobile (1.000) / PandaGripper (0.905) \\ Stovetop (0.965) / PandaMobile (0.893) \\ Wall (0.941) / Dishwasher (0.813) } & \shortstack{Stove (1.000) \\ PandaMobile (1.000) \\ PandaGripper (0.999) \\ Hood (0.168) / OmronMobileBase (0.216) \\ Floor (0.123) / SingleCabinet (0.167) } & \shortstack{Fridge (1.000) / PandaMobile (0.994) \\ PandaMobile (0.999) / PandaGripper (0.846) \\ PandaGripper (0.912) / Toaster (0.701) \\ Microwave (0.641) / WallAccessory (0.598) \\ Stove (0.572) } \\
\hline
TurnOffStove & \shortstack{Microwave (1.000) / Stove (0.955) \\ PandaMobile (1.000) / Dishwasher (0.931) \\ Wall (0.999) / PandaGripper (0.927) \\ Stove (0.999) / PandaMobile (0.879) \\ Stovetop (0.952) / Fridge (0.785) } & \shortstack{Stove (1.000) / PandaMobile (1.000) \\ PandaMobile (1.000) / PandaGripper (1.000) \\ PandaGripper (0.993) / Stove (1.000) \\ Floor (0.159) / Stovetop (0.086) \\ Accessory (0.139) / WallAccessory (0.066) } & \shortstack{PandaMobile (0.991) \\ PandaGripper (0.928) \\ WallAccessory (0.583) / Sink (0.865) \\ Microwave (0.529) / Drawer (0.833) \\ Stove (0.493) / CoffeeMachine (0.587) } \\
\hline
\end{tabular}
}
\end{table*}

\begin{table*}[t]
\caption{Five most relevant nodes per task and model – Lever.}
\label{tab:FiveMostRelevantNodes_Lever}
\centering
\resizebox{\linewidth}{!}{
\begin{tabular}{l|l|l|l}
Task & TopK & SIR & Threshold \\
\hline
TurnOnSinkFaucet & \shortstack{PandaMobile (0.965) \\ FramedWindow (0.956) / Fridge (0.851) \\ Sink (0.888) \\ Wall (0.864) / distr\_counter\_1 (0.628) \\ Counter (0.673) / WallAccessory (0.586) } & \shortstack{Sink (1.000) \\ PandaMobile (0.999) \\ PandaGripper (0.995) \\ Stool (0.016) \\ Counter (0.002) / Wall (0.002) } & \shortstack{PandaMobile (0.998) \\ FramedWindow (0.839) / PandaGripper (0.607) \\ Toaster (0.780) / FramedWindow (0.389) \\ PandaGripper (0.735) / distr\_counter\_1 (0.331) \\ WallAccessory (0.546) / distr\_counter\_0 (0.321) } \\
\hline
TurnOffSinkFaucet & \shortstack{PandaMobile (0.995) \\ Sink (0.956) / distr\_counter\_0 (0.737) \\ FramedWindow (0.845) / distr\_counter\_1 (0.731) \\ Counter (0.809) / PandaGripper (0.686) \\ Stool (0.797) / Sink (0.658) } & \shortstack{PandaMobile (1.000) \\ Sink (1.000) \\ PandaGripper (0.999) \\ HingeCabinet (0.002) / Stool (0.005) \\ Wall (0.000) / Floor (0.002) } & \shortstack{PandaMobile (1.000) / Toaster (1.000) \\ FramedWindow (0.971) / PandaMobile (1.000) \\ Fridge (0.919) / PandaGripper (0.704) \\ PandaGripper (0.610) / FramedWindow (0.560) \\ WallAccessory (0.508) / Fridge (0.527) } \\
\hline
TurnSinkSpout & \shortstack{PandaMobile (0.990) \\ Sink (0.960) / distr\_counter\_0 (0.838) \\ Wall (0.924) / PandaGripper (0.809) \\ FramedWindow (0.879) / distr\_sink (0.666) \\ Counter (0.877) / distr\_counter\_2 (0.635) } & \shortstack{Sink (1.000) / PandaMobile (1.000) \\ PandaMobile (1.000) / Sink (0.995) \\ PandaGripper (0.976) \\ Counter (0.001) / Floor (0.056) \\ Wall (0.001) / Counter (0.000) } & \shortstack{PandaMobile (1.000) \\ FramedWindow (0.911) / Fridge (1.000) \\ PandaGripper (0.884) \\ Sink (0.559) / FramedWindow (0.841) \\ distr\_counter\_0 (0.527) / WallAccessory (0.651) } \\
\hline
\end{tabular}
}
\end{table*}

\begin{table*}[t]
\caption{Five most relevant nodes per task and model – Button.}
\label{tab:FiveMostRelevantNodes_Button}
\centering
\resizebox{\linewidth}{!}{
\begin{tabular}{l|l|l|l}
Task & TopK & SIR & Threshold \\
\hline
CoffeePressButton & \shortstack{PandaMobile (0.988) \\ Counter (0.974) / PandaGripper (0.941) \\ Wall (0.897) / obj (0.901) \\ PandaGripper (0.857) / CoffeeMachine (0.810) \\ Fridge (0.741) / Toaster (0.724) } & \shortstack{PandaMobile (1.000) \\ PandaGripper (0.995) \\ CoffeeMachine (0.992) \\ obj (0.008) \\ WallAccessory (0.000) / Wall (0.007) } & \shortstack{PandaMobile (0.992) \\ PandaGripper (0.930) / FramedWindow (0.961) \\ Sink (0.820) / PandaGripper (0.838) \\ obj (0.522) / CoffeeMachine (0.667) \\ WallAccessory (0.495) / Toaster (0.644) } \\
\hline
TurnOnMicrowave & \shortstack{PandaMobile (0.995) / Toaster (1.000) \\ Microwave (0.992) / PandaGripper (0.978) \\ Oven (0.983) / PandaMobile (0.754) \\ PandaGripper (0.678) / Oven (0.728) \\ Wall (0.607) / OmronMobileBase (0.717) } & \shortstack{PandaGripper (0.997) / Microwave (1.000) \\ PandaMobile (0.997) \\ Microwave (0.994) / PandaGripper (0.997) \\ Oven (0.014) / HingeCabinet (0.029) \\ Counter (0.005) } & \shortstack{PandaMobile (0.985) \\ Toaster (0.940) / PandaGripper (0.619) \\ PandaGripper (0.776) / Microwave (0.222) \\ Oven (0.425) / OmronMobileBase (0.201) \\ Microwave (0.249) / CoffeeMachine (0.163) } \\
\hline
TurnOffMicrowave & \shortstack{Microwave (1.000) / Toaster (0.960) \\ PandaMobile (0.999) / Microwave (0.872) \\ Oven (0.991) / PandaGripper (0.841) \\ PandaGripper (0.794) / Oven (0.815) \\ Wall (0.622) / PandaMobile (0.713) } & \shortstack{PandaGripper (1.000) / PandaMobile (1.000) \\ Microwave (1.000) / PandaGripper (0.997) \\ PandaMobile (1.000) / Microwave (0.996) \\ Counter (0.000) / HingeCabinet (0.064) \\ WallAccessory (0.000) } & \shortstack{PandaMobile (1.000) \\ PandaGripper (0.870) \\ Toaster (0.768) / Microwave (0.256) \\ Oven (0.493) / OmronMobileBase (0.176) \\ Microwave (0.380) / Floor (0.132) } \\
\hline
\end{tabular}
}
\end{table*}

\begin{table*}[t]
\caption{Five most relevant nodes per task and model – Insertion.}
\label{tab:FiveMostRelevantNodes_Insertion}
\centering
\resizebox{\linewidth}{!}{
\begin{tabular}{l|l|l|l}
Task & TopK & SIR & Threshold \\
\hline
CoffeeServeMug & \shortstack{PandaMobile (0.997) / CoffeeMachine (0.967) \\ Wall (0.981) / PandaMobile (0.951) \\ Counter (0.963) / obj (0.895) \\ Sink (0.760) / PandaGripper (0.848) \\ PandaGripper (0.744) / Toaster (0.760) } & \shortstack{obj (1.000) / CoffeeMachine (1.000) \\ CoffeeMachine (1.000) / Counter (1.000) \\ Counter (0.999) / PandaMobile (1.000) \\ PandaMobile (0.995) / obj (0.999) \\ PandaGripper (0.981) } & \shortstack{PandaMobile (0.994) \\ Sink (0.924) / FramedWindow (0.998) \\ PandaGripper (0.921) / CoffeeMachine (0.958) \\ CoffeeMachine (0.846) / PandaGripper (0.935) \\ WallAccessory (0.695) / Sink (0.872) } \\
\hline
CoffeeSetupMug & \shortstack{FramedWindow (0.994) / PandaMobile (0.991) \\ PandaMobile (0.992) / obj (0.989) \\ Wall (0.981) / CoffeeMachine (0.964) \\ Counter (0.969) / PandaGripper (0.917) \\ PandaGripper (0.883) / Toaster (0.722) } & \shortstack{Counter (1.000) / PandaMobile (1.000) \\ CoffeeMachine (0.999) / Counter (1.000) \\ PandaGripper (0.980) / CoffeeMachine (1.000) \\ PandaMobile (0.974) / obj (0.995) \\ obj (0.964) / PandaGripper (0.978) } & \shortstack{FramedWindow (1.000) / PandaMobile (1.000) \\ PandaMobile (0.993) / PandaGripper (0.991) \\ PandaGripper (0.978) / CoffeeMachine (0.991) \\ CoffeeMachine (0.797) / Toaster (0.973) \\ WallAccessory (0.692) / Sink (0.655) } \\
\hline
\end{tabular}
}
\end{table*}

\subsection{Grad-CAM Visualizations}
\begin{figure*}[t]
\centering
\setlength{\tabcolsep}{2pt}
\renewcommand{\arraystretch}{0}

\begin{tabular}{cc}
\includegraphics[width=0.31\textwidth]{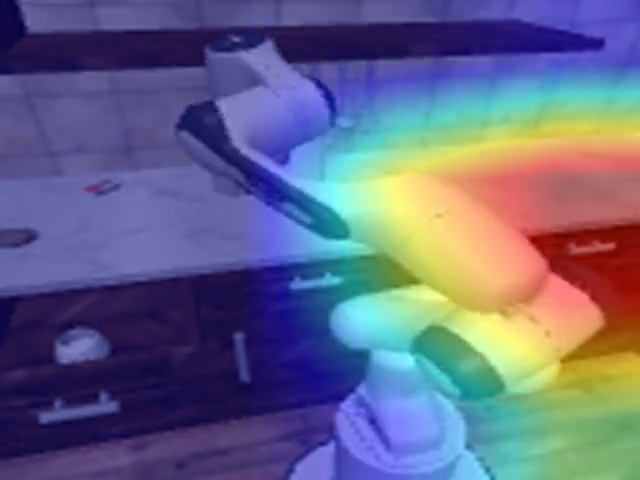}  &
\includegraphics[width=0.31\textwidth]{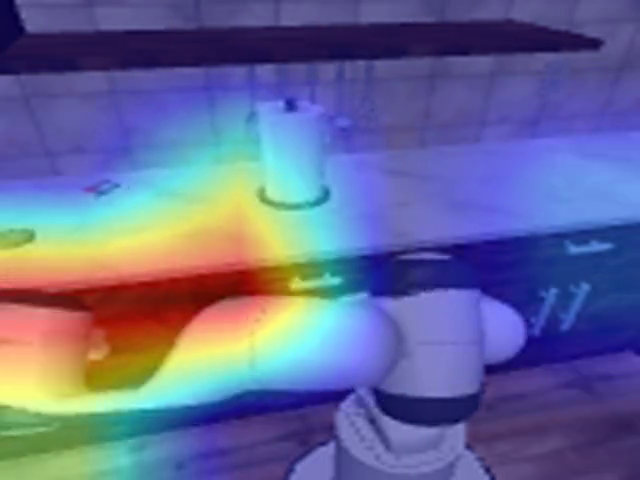} \\[3pt]

\includegraphics[width=0.31\textwidth]{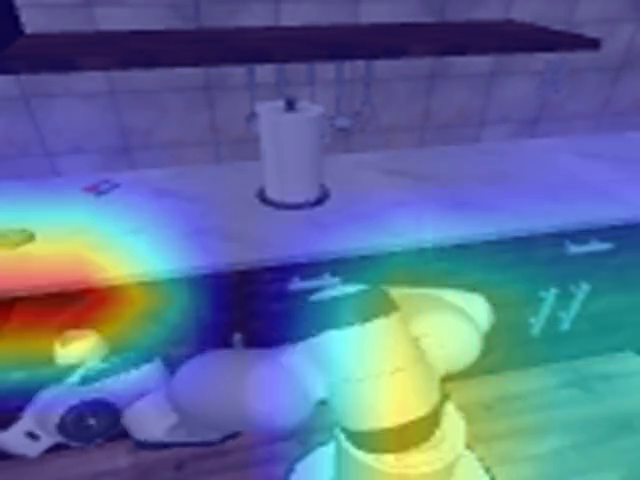} &
\includegraphics[width=0.31\textwidth]{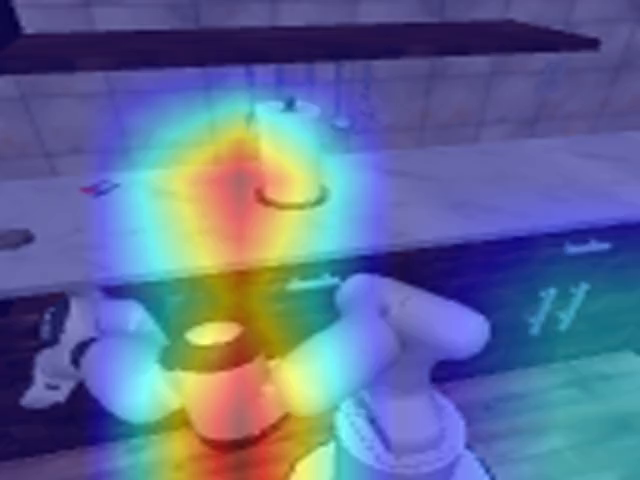} \\

\end{tabular}

\caption{Grad-CAM explanations over time for a rollout of the \textit{CloseDrawer} task (frames 20, 70, 120, 170). 
We find that the explanations produced by Grad-CAM are neither consistent, as the activated regions are jumping around a lot in the image, nor expressive, as the highlighted regions are not bound to objects.}
\label{fig:GradCAM_CloseDrawer}
\end{figure*}

We visualize four steps of a rollout using the image baseline on the \textit{CloseDrawer} task in \Cref{fig:GradCAM_CloseDrawer}.
The model's focus shifts between the robot arm, gripper, and counter/drawer.
Over the whole rollout the focus includes nearly the entire image. 
This lack of precision renders such visualizations ineffective for interpreting model behaviour or gaining insights into the dataset.

\end{document}